\documentclass[10pt,twocolumn,letterpaper]{article}

\usepackage{cvpr}
\usepackage{times}
\usepackage{epsfig}
\usepackage{graphicx}
\usepackage{amsmath}
\usepackage{amssymb}
\usepackage{tikz}
\usepackage{subcaption}
\usetikzlibrary{positioning}
\usepackage{multirow}
\usepackage{dsfont}
\usepackage[font={small}]{caption}
\usepackage[export]{adjustbox}
\usepackage{epstopdf}
\DeclareGraphicsExtensions{.png,.pdf}
\usetikzlibrary{shapes.callouts}
\usetikzlibrary{arrows}
\usetikzlibrary{chains, fit, quotes, automata, arrows}
\usetikzlibrary{calc}
\usepackage{mathtools}
\usepackage{booktabs}
\usepackage{array}
\usepackage{footmisc}
\usepackage{xcolor}
\usepackage{enumitem}

\newcolumntype{M}[1]{>{\centering\arraybackslash}m{#1}}
\DeclareMathOperator*{\argmin}{arg\,min}
\DeclareMathOperator*{\argmax}{arg\,max}

\definecolor{c1}{RGB}{38,17,191}
\definecolor{c5}{RGB}{29,122,213}
\definecolor{c3}{RGB}{77,175,74}
\definecolor{c4}{RGB}{154,87,142}
\definecolor{c2}{RGB}{255,127,0}

\tikzset{
  level/.style   = { ultra thick, black },
  connect/.style = { dashed, red },
  notice/.style  = { draw, rectangle callout, callout relative pointer={#1} },
  label/.style   = { text width=2cm }
}

\definecolor{purp}{RGB}{112, 48, 160}

\usepackage[pagebackref=true,breaklinks=true,letterpaper=true,colorlinks,bookmarks=false]{hyperref}

\cvprfinalcopy

\ifcvprfinal\fi

\makeatletter
\g@addto@macro \normalsize {%
 \setlength\abovedisplayskip{5pt plus 2pt minus 2pt}%
 \setlength\belowdisplayskip{5pt plus 2pt minus 2pt}%
}
\makeatother

\begin{document}

\title{Identifying Most Walkable Direction for Navigation in an Outdoor Environment}

\author{Sachin Mehta, Hannaneh Hajishirzi, and Linda Shapiro\\
University of Washington, Seattle\\
Email: \{sacmehta, hannaneh\}@uw.edu, shapiro@cs.washington.edu}

\maketitle

\begin{abstract}
We present an approach for identifying the most walkable direction for navigation using a hand-held camera. Our
approach extracts semantically rich contextual information from the scene using a custom encoder-decoder architecture for semantic segmentation and models the spatial and temporal behavior of objects in the scene using a spatio-temporal graph. The system learns to minimize a cost function over the spatial and temporal object attributes to identify the most walkable direction. We construct a  new annotated navigation dataset collected using a hand-held mobile camera in an unconstrained outdoor environment, which includes challenging settings such as highly dynamic scenes, occlusion between objects, and distortions. Our system achieves an accuracy of  $84\%$ on predicting a safe direction. We also show that our custom segmentation network is both fast and accurate, achieving mIOU (mean intersection over union) scores of 81 and 44.7 on the PASCAL VOC and the PASCAL Context datasets, respectively, while running at about 21 frames per second. 
\end{abstract}
\vspace{-2.0mm}
\section{Introduction}
\vspace{-2.0mm}
White canes and guide dogs are the two most popular mobility aids used by the blind. However, only 2\% use guide dogs \cite{nfb2016}, primarily due to behavioral issues (such as fear, anxiety, and aggression) related to aging dogs \cite{caron2016using}. The most widely used mobility aid, white canes, rely on contact with obstacles, as a source of feedback and therefore, pose collision risks. To address these risks, systems that use sensing devices, such as ultrasonic transducers \cite{ran2004drishti}, RFID tags \cite{bessho2008assisting}, and cameras \cite{apostolopoulos2014integrated, lee2016rgb, li2016isana, phung2016pedestrian} have been proposed. These systems have proven to be robust and accurate for \textit{indoor} navigation. Systems that use affordable RGB-D cameras are particularly effective as they provide rich information about the surrounding environment \cite{lee2016rgb, li2016isana, aladren2016navigation}. However, these systems cannot be effectively deployed in an unconstrained outdoor environment due to challenges that include highly dynamic scenes, uneven road surfaces, occlusions, and most importantly, poor performance of RGB-D cameras in outdoor environments. 

\begin{figure}[t!]
\centering
\begin{subfigure}[b]{0.4\columnwidth}
\centering
\includegraphics[width=\columnwidth]{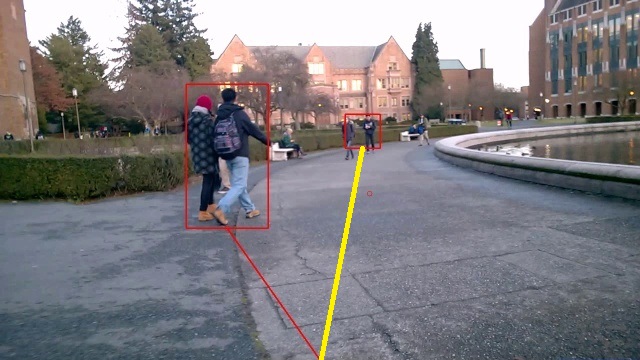}
\caption{Frame 1 (Slight right)}
\end{subfigure}
\hfill
\begin{subfigure}[b]{0.4\columnwidth}
\centering
\includegraphics[width=\columnwidth]{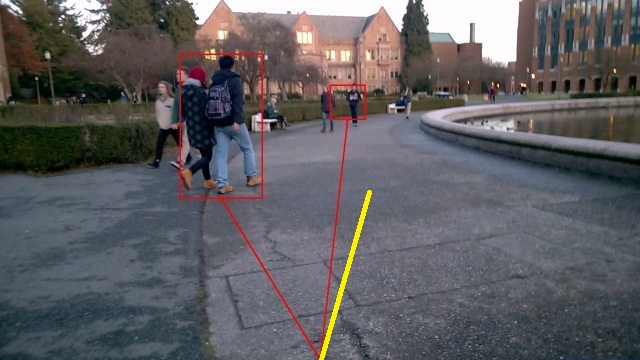}
\caption{Frame 31 (Slight right)}
\end{subfigure}
\vfill
\begin{subfigure}[b]{0.4\columnwidth}
\centering
\includegraphics[width=\columnwidth]{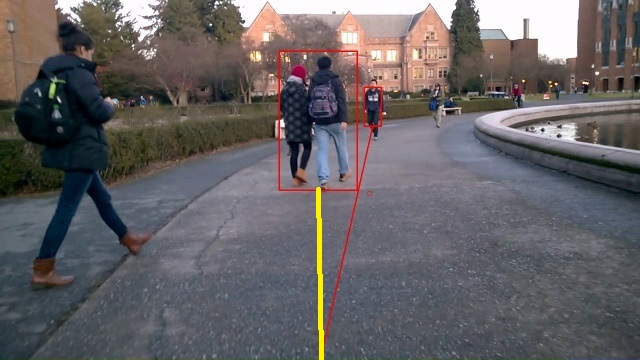}
\caption{Frame 81 (Straight)}
\end{subfigure}
\hfill
\begin{subfigure}[b]{0.4\columnwidth}
\centering
\includegraphics[width=\columnwidth]{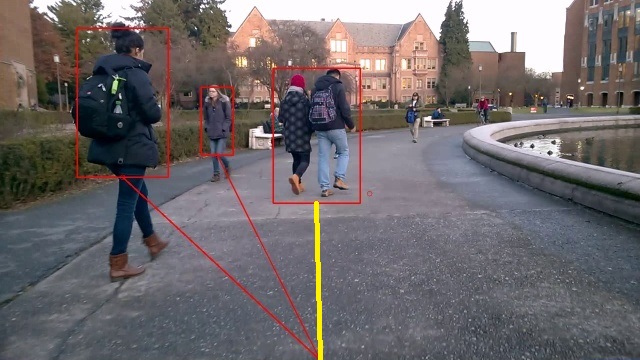}
\caption{Frame 110 (Straight)}
\end{subfigure}
\setlength{\belowcaptionskip}{-3mm}
\caption{Identifying most walkable directions. Without any temporal information, our method identified the farthest object node as the safest node (a). Over time, our method found that the object is approaching the camera and therefore, it updated the walkable node to the ground object (b). After observing the environment for about 4 seconds, our method identified an object that is moving in the same direction as the camera and is \textit{safe} (c, d). The yellow line denotes the safest detected node or object; the generated navigational cues are written in the sub-caption in brackets.}
\label{fig:exampleNav}
\end{figure}

In this work, we are interested in identifying the most walkable directions that could help visually impaired people  to navigate safely in an unconstrained outdoor environment using a \textit{hand-held camera} device. The \textit{most walkable direction} is the direction corresponding to a safe-to-follow object that could help minimize collisions with objects while helping in navigation. A hand-held camera offers a first-person view that cannot be captured using location sensors such as beacons and GPS. For example, an intelligent system can identify a person (at some safe distance) and provide navigational directions that could help the navigator to follow that person. This could help blind people to avoid head-high collisions that are difficult to avoid with a white cane, the most popular mobility aid among the blind. An example generated from our method is shown in Figure \ref{fig:exampleNav}.

\begin{figure*}[ht!]
\centering
\begin{subfigure}[b]{0.3\columnwidth}
\includegraphics[width=\columnwidth]{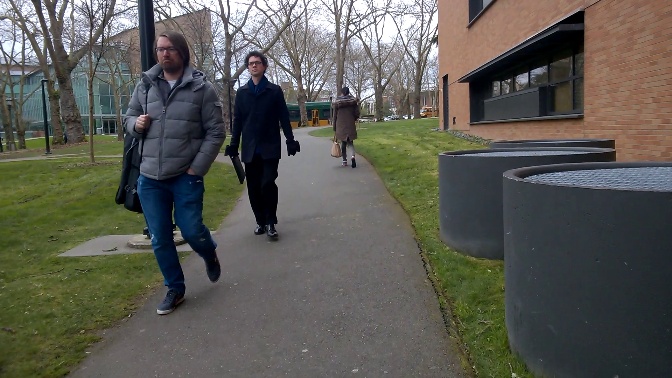}
\caption{}
\end{subfigure}
\hfill
\begin{subfigure}[b]{0.3\columnwidth}
\includegraphics[width=\columnwidth]{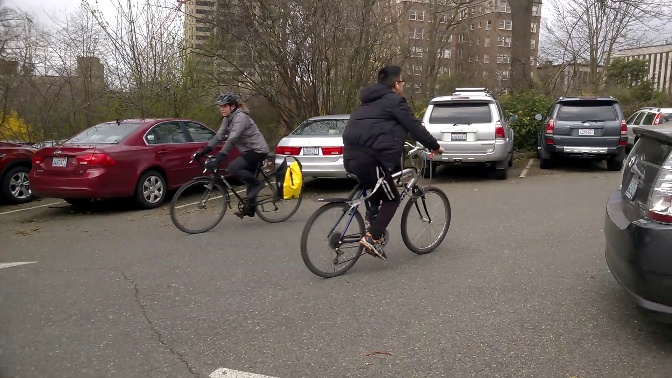}
\caption{}
\end{subfigure}
\hfill
\begin{subfigure}[b]{0.3\columnwidth}
\includegraphics[width=\columnwidth]{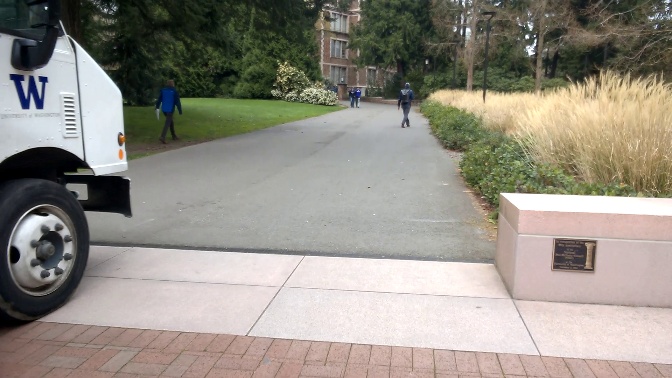}
\caption{}
\end{subfigure}
\hfill
\begin{subfigure}[b]{0.3\columnwidth}
\includegraphics[width=\columnwidth]{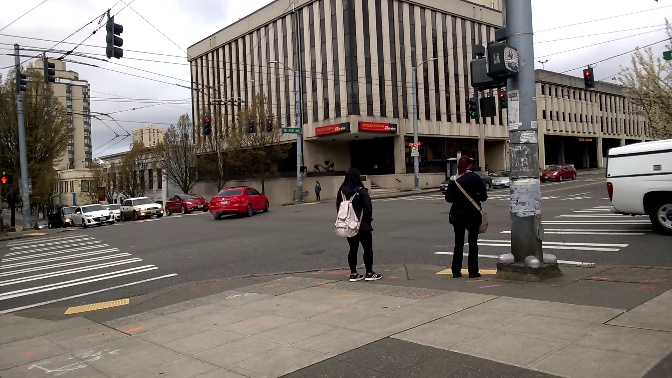}
\caption{}
\end{subfigure}
\hfill
\begin{subfigure}[b]{0.3\columnwidth}
\includegraphics[width=\columnwidth]{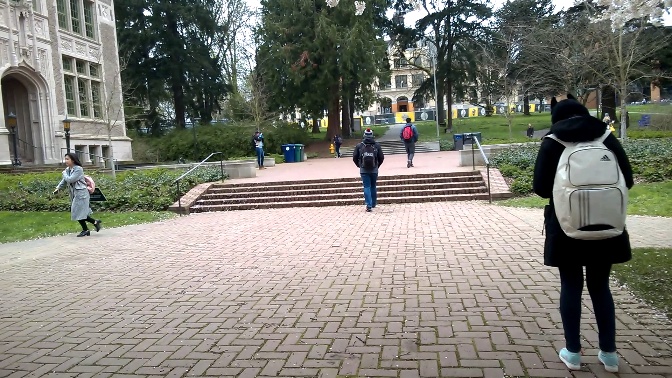}
\caption{}
\end{subfigure}
\hfill
\begin{subfigure}[b]{0.3\columnwidth}
\includegraphics[width=\columnwidth, height=40px]{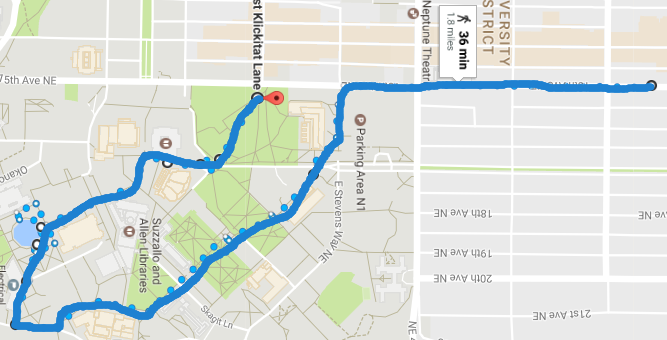}
\caption{}
\label{fig:areamap}
\end{subfigure}
\vfill
\begin{subfigure}[b]{0.3\columnwidth}
\includegraphics[width=\columnwidth]{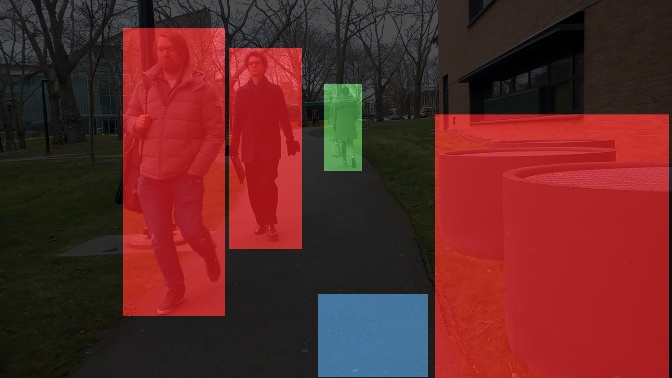}
\caption{}
\end{subfigure}
\hfill
\begin{subfigure}[b]{0.3\columnwidth}
\includegraphics[width=\columnwidth]{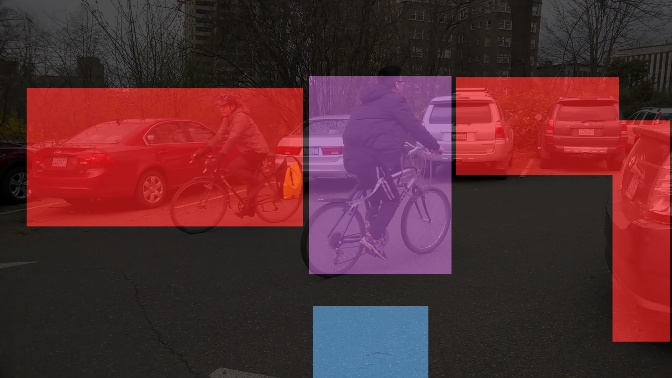}
\caption{}
\end{subfigure}
\hfill
\begin{subfigure}[b]{0.3\columnwidth}
\includegraphics[width=\columnwidth]{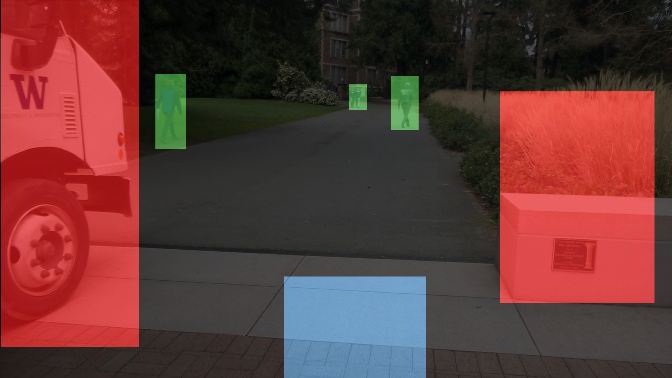}
\caption{}
\end{subfigure}
\hfill
\begin{subfigure}[b]{0.3\columnwidth}
\includegraphics[width=\columnwidth]{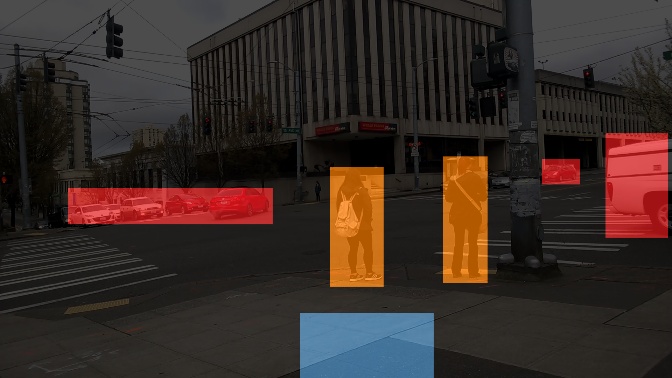}
\caption{}
\end{subfigure}
\hfill
\begin{subfigure}[b]{0.3\columnwidth}
\includegraphics[width=\columnwidth]{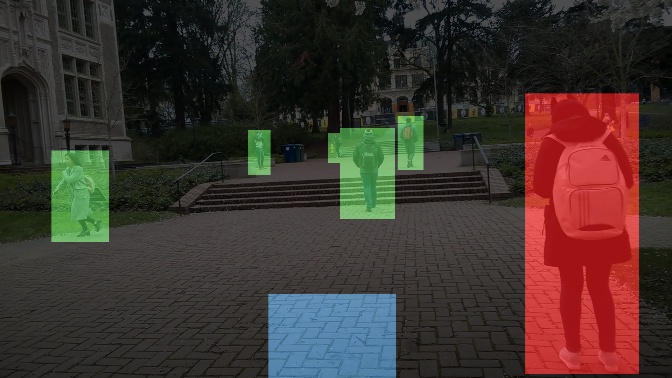}
\caption{}
\end{subfigure}
\hfill
\begin{subfigure}[b]{0.35\columnwidth}
\resizebox{\columnwidth}{!}{
\definecolor{g}{RGB}{55 ,126, 184}
\definecolor{ah}{RGB}{228, 26, 28}
\definecolor{am}{RGB}{255,127,0}
\definecolor{al}{RGB}{152,78,163}
\definecolor{sa}{RGB}{77,175, 77}

\begin{tikzpicture}[
block/.style={
align=center,rounded corners,
font=\Large}]
\node [draw,thick,minimum width=0.75cm,minimum height=0.75cm, fill=g] (gr) {};
\node[right=0.25cm of gr, align=left] (gr_t) {\LARGE Best safe area on the ground};

\node [draw,thick,minimum width=0.75cm,minimum height=0.75cm, fill=ah, below=0.25cm of gr] (ah) {};
\node[right=0.25cm of ah, align=left] (ah_t) {\LARGE Safety level of object: Avoid High};

\node [draw,thick,minimum width=0.75cm,minimum height=0.75cm, fill=am, below=0.25cm of ah] (am) {};
\node[right=0.25cm of am, align=left] (am_t) {\LARGE Safety level of object: Avoid Medium};

\node [draw,thick,minimum width=0.75cm,minimum height=0.75cm, fill=al, below=0.25cm of am] (al) {};
\node[right=0.25cm of al, align=left] (al_t) {\LARGE Safety level of object: Avoid Low};

\node [draw,thick,minimum width=0.75cm,minimum height=0.75cm, fill=sa, below=0.25cm of al] (sf) {};
\node[right=0.25cm of sf, align=left] (sf_t) {\LARGE Safety level of object: Safe};

\end{tikzpicture}
}
\caption{}
\end{subfigure}
\setlength{\belowcaptionskip}{-4mm}
\caption{Samples from our dataset. RGB images (a-e) along with their corresponding annotations (g-k). Our dataset was captured over an area of 1.8 miles, a walking distance of approx. 40 minutes for a normal sighted person (f). Our dataset is challenging due to several factors such as different ground surfaces (e.g. concrete (a,b), cement (d), and tiling (c, e)), different lighting conditions (a, b, e), random appearance of the objects (e.g. truck appearing in (c)), occlusions (b), and different type of roads (narrow (a), wider (e), and road intersections(d)).}
\label{fig:dataset}
\end{figure*}
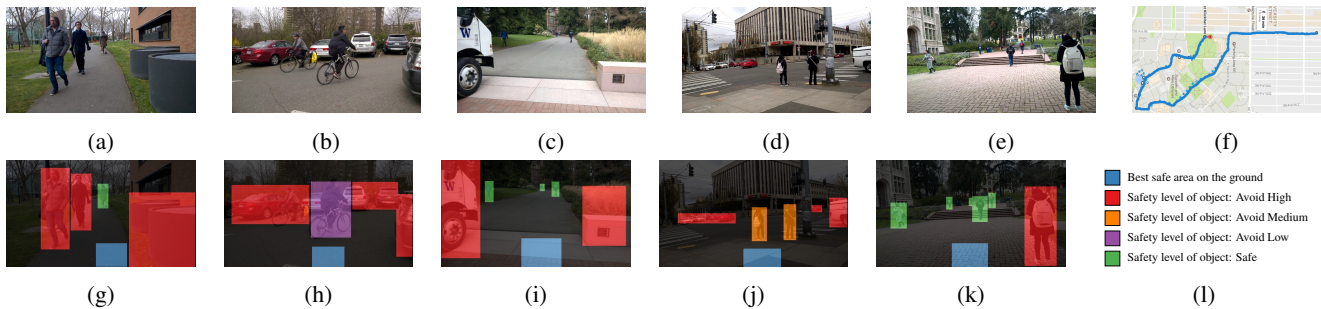

An outdoor environment is highly dynamic, and objects (ambulatory as well as non-ambulatory) appear randomly at random times. Often there are no ambulatory and non-ambulatory objects. Therefore, a system for identifying the most walkable direction should be able to adapt to the changes in the outdoor environment and should be independent of any specific object category (e.g. person). Motivated by the challenges that these unconstrained outdoor environments present in navigation, we propose an approach that learns the \textit{category-aware cues} to identify the most walkable direction. Our work represents  semantically segmented videos using spatio-temporal graphs, leveraging a fast and accurate semantic segmentation algorithm. Our method learns to minimize a cost function from the spatial and temporal attributes of the objects to identify the most walkable direction present in the scene.

The main contributions of our paper are:
(1) A novel approach for identifying the most walkable directions in the scene by learning cost functions over spatial and temporal attributes, which could aid in navigation in an unconstrained outdoor environment while avoiding collisions using a hand-held device. 
(2) Applications that aim at navigation demand accurate predictions with low-latency. Our experiments show that existing methods are either accurate or fast; but not both. We use prior knowledge in the domain of CNNs and propose a custom architecture that is fast as well as accurate. Our method achieves mIOU (mean intersection over union) scores of 81 and 44.7 on the PASCAL VOC and the PASCAL Context datasets, respectively, while being much faster than the state-of-the-art methods\footnote{Source code is available at: \url{https://github.com/sacmehta/MSRSegNet}}.
(3) We create a new annotated dataset comprising 40,000+ images. Our dataset is specifically designed to facilitate the development of robust assistive technologies for navigation in unconstrained outdoor environments (Figure \ref{fig:dataset}). Our method achieves an accuracy of $84\%$ on this dataset.
\vspace{-2.0mm}
\section{Related Work}
\vspace{-2.0mm}
Previous work in HCI community use crowd-sourcing \cite{bigham2010vizwiz} or direct sensing devices -- such as RFID tags \cite{ran2004drishti}, ultrasonic sensors \cite{bessho2008assisting}, and mobile phone sensors \cite{brock2015interactivity, guy2012crossingguard, link2011footpath} -- to assist the blind. However, these approaches do not work in unconstrained outdoor environments.

Visual simultaneous localization and mapping (SLAM) methods have been widely used in robotics to localize a robot's position and construct a map using cameras \cite{kerl2013dense, engel2014lsd, mur2015orb, zhang2015building}. Feature-based methods \cite{forster2014svo} have been complemented with region-based approaches \cite{concha2014using}, direct visual odometry methods\footnote{Direct visual odometry uses image information to construct depth maps.} \cite{forster2014svo}, pose graph-based optimization techniques \cite{kerl2013dense, liu2011robust}, and GPS \cite{brilhault2011fusion}. SLAM-based approaches for assistive technologies \cite{apostolopoulos2014integrated, fallah2012user, aladren2016navigation, pradeep2010robot, hardegger2013actionslam} are difficult to scale in an unconstrained outdoor environment because they: (1) make assumptions about the environment \cite{apostolopoulos2014integrated, fallah2012user} and vicinity of the object \cite{pradeep2010robot}, and (2) use devices suitable for indoor environments \cite{aladren2016navigation}. 

Systems based on detection and tracking methods have been proposed for assistive technologies. Marker-based methods use simple detection algorithms (such as edge detection) to detect and locate specific markers installed in the environment, with \cite{li2016isana} or without \cite{manduchi2010blind} pre-existing environmental maps. Marker-less approaches use more sophisticated algorithms, such as SIFT and OCR \cite{pradeep2008piecewise, phung2016pedestrian, liu2015isee,nicholson2009shoptalk}. 

Spatio-temporal graphs have been used in applications, such as video summarization \cite{lee2012discovering}, driver assistance \cite{zhang2014overtaking, jain2015car} and activity recognition \cite{brendel2011learning,koppula2016anticipating,jain2016structural}, for reasoning about spatial and temporal relationships between objects. We extend the previous work on spatio-temporal graphs for identifying the most walkable directions in the scene. We extract semantically rich contextual information from the scene using semantic segmentation. Most current semantic segmentation networks such as RefineNet \cite{lin2016refinenet} and DeepLab-v2 \cite{chen2016deeplab} are accurate but slow. We extend the previous work on semantic segmentation and improves on them with carefully designed components that makes our network fast while delivering state-of-the-art segmentation results.
\vspace{-2.0mm}
\section{Overview}
\vspace{-2.0mm}
\label{sec:ourMethod}
In this paper, our goal is to identify the most walkable directions in order to help visually impaired people navigate safely in an outdoor environment using a \textit{hand-held camera}. We identify the most walkable direction by finding an object that is safe to follow, while avoiding collisions. 
We use a spatio-temporal graph to represent interactions between the objects and the camera over time. The system learns a cost function over spatial and temporal attributes of objects and minimizes the cost to find the most walkable direction. The training data includes annotations about the safety-level of the objects (safe to follow or avoid) along with the best walkable surface that is safe to walk in the next time step. Several samples from our dataset with their annotations are visualized in Figure \ref{fig:dataset}. 

\begin{figure}[t!]
\setlength{\belowcaptionskip}{-4mm}
\centering
\includegraphics[width=0.9\columnwidth]{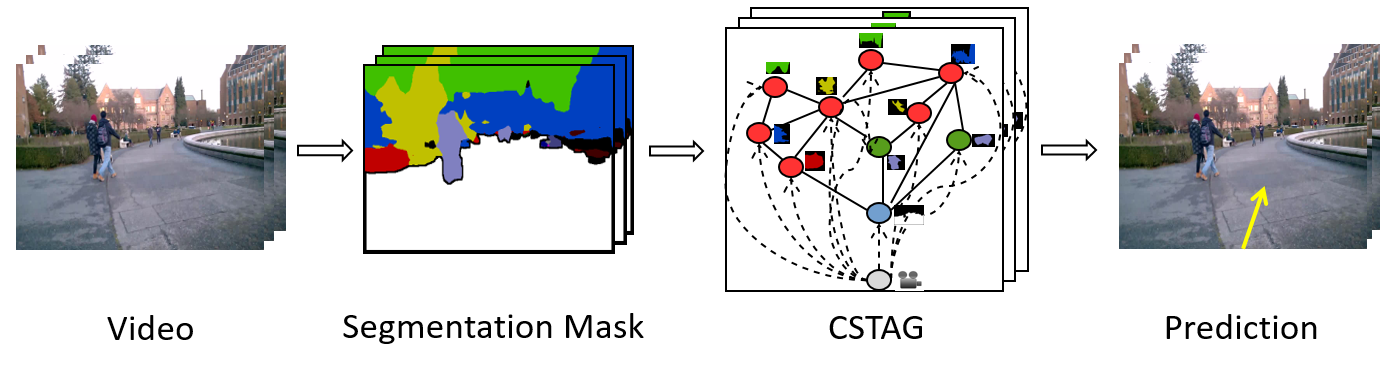}
\caption{Overview of our approach. Videos are segmented and represented by category-aware spatio-temporal attribute graphs (CSTAGs). These CSTAGs are then used to learn the cost function that describes the relationship between objects and camera over time and helps in determining the most walkable direction while avoiding collisions.}
\label{fig:overview}
\end{figure}

An overview of our approach, consisting of the following steps, is shown in Figure \ref{fig:overview}.  (1) It starts with a semantic segmentation of each frame; (2) It represents the segmented objects as a spatio-temporal graph and calculates attributes that can describe the behavior of the objects. (3) It then learns  a cost function over these attributes to identify the most walkable direction. We first describe the representation (Section \ref{ssec:represntation}) and the learning algorithm (Section \ref{ssec:costfunction}) and then describe our segmentation method (Section \ref{sec:semanticSeg}) which is fast, accurate, and suitable for navigating guidelines.

\vspace{-2.0mm}
\section{Representation: Spatio-Temporal Graph }
\label{ssec:represntation}
\begin{figure}[b!]
\centering\begin{subfigure}[b]{0.45\columnwidth}
\centering
\resizebox{0.88\columnwidth}{!}{
\input{tikz/category.tikz}
}
\caption{Object categories}
\label{fig:magConstruct}
\end{subfigure}
\hfill
\begin{subfigure}[b]{0.45\columnwidth}
\centering
\includegraphics[width=0.9\columnwidth]{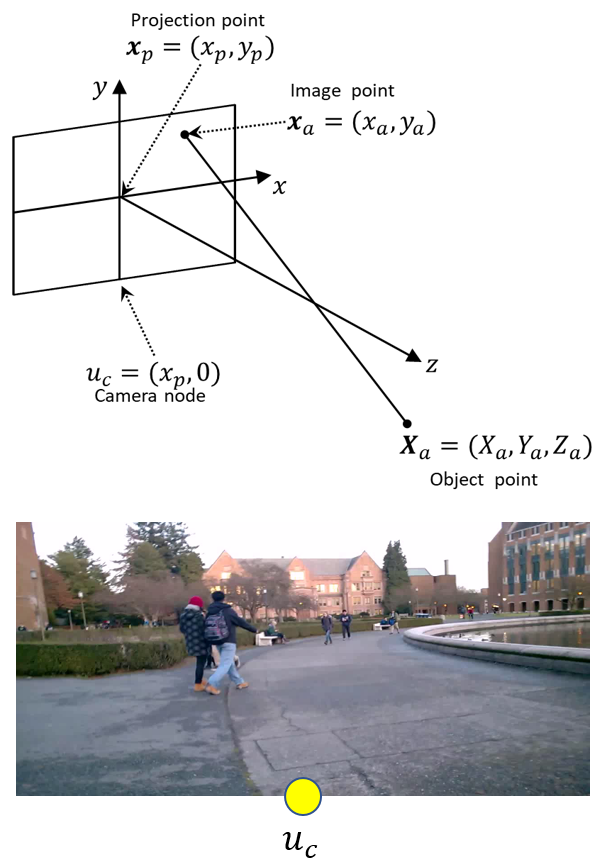}
\caption{Camera node $u_c$}
\label{fig:camNode}
\end{subfigure}
\setlength{\belowcaptionskip}{-2mm}
\caption{Visualization of object categorization (a) and camera node (b). Camera node $u_c$ is the projection of principal point onto the horizontal and is marked in \textbf{yellow} in (a). Here, $X_A$ is an object point in 3-D space, which is projected onto the image plane to create an image point $x_a$. \{$X$,$Y$,$Z$\} and \{$x$,$y$\} are the real-world and camera coordinate systems. The principal or camera axis (denoted by $z$-axis) is parallel to the horizontal.}
\label{fig:camNodeA}
\end{figure}

\vspace{-2.0mm}
Spatio-temporal graphs enable reasoning about spatio-temporal activity \cite{lee2012discovering,koppula2016anticipating,jain2016structural} and keep track of the objects in the scene, allowing our system to avoid running a full segmentation at every frame of the video. We represent a segmented video sequence by a spatio-temporal graph called a category-aware spatio-temporal attribute graph (CSTAG).  A CSTAG is a spatio-temporal graph $\mathcal{G} = (\mathcal{U}, \mathcal{E}_S, \mathcal{E}_T)$, where $\mathcal{U}$ is a set of attributed nodes, $\mathcal{E}_S$ is a set of attributed spatial edges, and $\mathcal{E}_T$ is a set of attributed temporal edges and whose structure $(\mathcal{U}, \mathcal{E}_S)$ unrolls over time through the edges $\mathcal{E}_T$. The set  $\mathcal{U} = \{ u_c\} \cup \mathcal{U}_o$ represents the nodes in the graph corresponding to the camera $u_c$ and objects $\mathcal{U}_o$ present in the image, which are grouped into three categories: (1) ground, (2) ambulatory objects, and (3) non-ambulatory objects (Figure \ref{fig:magConstruct}). The camera node $u_c$ represents the projection of the principal point onto the horizontal, as shown in Figure \ref{fig:camNode}, while the spatial locations of nodes in $\mathcal{U}_o$ are represented by their centroids in a graph.

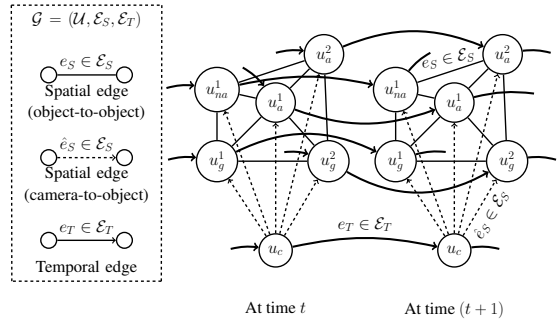
\begin{figure}[b!]
\setlength{\belowcaptionskip}{-4mm}
\centering
\resizebox{0.9\columnwidth}{!}{\begin{tikzpicture}  
  \LARGE{
  	\node[circle, draw, fill=white, line width=0.5mm] (u) {$u_a^1$};
    \node[circle, draw, fill=white, below left = 1.5cm of u, line width=0.5mm] (v) {$u_g^1$};
    \node[circle, draw, fill=white, right = 2.5cm of v, line width=0.5mm] (g1) {$u_g^2$};
    \node[circle, draw, fill=white, above left = 1cm of u, above= 1cm of v, line width=0.5mm] (w) {$u_{na}^1$};
    \node[circle, draw, fill=white, above right = 1cm of u, line width=0.5mm] (y) {$u_{a}^2$};
    \node[circle, draw, fill=white, below = 4cm of u, line width=0.5mm] (c) {$u_{c}$};
    \node[rectangle, fill=white, below = 1cm of c, line width=0.5mm] (desc) {At time $t$};

    \node[circle, draw, fill=white, line width=0.5mm, right= 5cm of u] (uA) {$u_a^1$};
    \node[circle, draw, fill=white, below left = 1.5cm of uA, line width=0.5mm] (vA) {$u_g^1$};
    \node[circle, draw, fill=white, right = 2.5cm of vA, line width=0.5mm] (g1A) {$u_g^2$};
    \node[circle, draw, fill=white, above left = 1cm of uA, above= 1cm of vA, line width=0.5mm] (wA) {$u_{na}^1$};
    \node[circle, draw, fill=white, above right = 1cm of uA, line width=0.5mm] (yA) {$u_{a}^2$};
    \node[circle, draw, fill=white, below = 4cm of uA, line width=0.5mm] (cA) {$u_{c}$};
    \node[rectangle,fill=white, below = 1cm of cA, line width=0.5mm] (desc1) {At time $(t+1)$};
    
    \node[circle, line width=0.5mm, above left= 1cm of u] (uA1) {};
    \node[circle, left= 1cm of v, line width=0.5mm] (vA1) {};
    \node[circle, below= 0.75cm of u, line width=0.5mm] (g1A1) {};
    \node[circle, left = 1cm of w, line width=0.5mm] (wA1) {};
    \node[circle, left = 1cm of y, line width=0.5mm] (yA1) {};
    \node[circle, left= 1cm of c, line width=0.5mm] (cA1) {};
    
    \node[circle, line width=0.5mm, below right= 1cm of yA] (uA2) {};
    \node[circle, below= 0.75cm of uA, line width=0.5mm] (vA2) {};
    \node[circle, right= 1cm of g1A, line width=0.5mm] (g1A2) {};
    \node[circle, above right = 1cm of wA, line width=0.5mm] (wA2) {};
    \node[circle, right = 1cm of yA, line width=0.5mm] (yA2) {};
    \node[circle, right= 1cm of cA, line width=0.5mm] (cA2) {};

    \draw[thick, solid, line width=0.5mm] (u) -- (w);
    \draw[thick, solid, line width=0.5mm] (w) -- (v);
    \draw[thick, solid, line width=0.5mm] (u) -- (v);
    
    \draw[thick, solid, line width=0.5mm] (w) --  (y);
    \draw[thick, solid, line width=0.5mm] (y) -- (u);
    \draw[thick, solid, line width=0.5mm] (v) -- (g1);
    \draw[thick, solid, line width=0.5mm] (g1) -- (u);
    \draw[thick, solid, line width=0.5mm] (g1) -- (y);
    
    \draw[thick, dashed, ->, line width=0.5mm] (c) -- (u);
    \draw[thick, dashed, ->, line width=0.5mm] (c) -- (v);
    \draw[thick, dashed, ->, line width=0.5mm] (c) -- (w);
    \draw[thick, dashed, ->, line width=0.5mm] (c) --  (g1);
    \draw[thick, dashed, ->, line width=0.5mm] (c) -- (y);

    \draw[thick, solid, line width=0.5mm] (uA) -- (wA);
    \draw[thick, solid, line width=0.5mm] (wA) -- (vA);
    \draw[thick, solid, line width=0.5mm] (uA) -- (vA);
    
    \draw[thick, solid, line width=0.5mm] (wA) -- node [pos=0.5,above,rotate=10] {$e_S \in \mathcal{E}_S$} (yA);
    \draw[thick, solid, line width=0.5mm] (yA) -- (uA);
    \draw[thick, solid, line width=0.5mm] (vA) -- (g1A);
    \draw[thick, solid, line width=0.5mm] (g1A) -- (uA);
    \draw[thick, solid, line width=0.5mm] (g1A) -- (yA);
    
    \draw[thick, dashed, ->, line width=0.5mm] (cA) -- (uA);
    \draw[thick, dashed, ->, line width=0.5mm] (cA) -- (vA);
    \draw[thick, dashed, ->, line width=0.5mm] (cA) -- (wA);
    \draw[thick, dashed, ->, line width=0.5mm] (cA) --  node [pos=0.5,below ,rotate=60] {$\hat{e}_S \in \mathcal{E}_S$} (g1A);
    \draw[thick, dashed, ->, line width=0.5mm] (cA) --  (yA);

    \draw[thick, ->, line width=0.75mm] (u) to [bend right=20] (uA);
    \draw[thick, ->, line width=0.75mm] (v) to [bend left=26] (vA);
    \draw[thick, ->, line width=0.75mm] (w) to [bend left=12] (wA);
    \draw[thick, ->, line width=0.75mm] (c) to [bend left=12] node [pos=0.5,above] {$e_T \in \mathcal{E}_T$}  (cA);
    \draw[thick, ->, line width=0.75mm] (g1) to [bend right=30] (g1A);
    \draw[thick, ->, line width=0.75mm] (y) to [bend left=22] (yA); 
    
     \draw[thick, ->, line width=0.75mm] (uA1) to [bend left=10] (u);
    \draw[thick, ->, line width=0.75mm] (vA1) to [bend left=10] (v);
    \draw[thick, ->, line width=0.75mm] (wA1) to [bend left=10] (w);
    \draw[thick, ->, line width=0.75mm] (cA1) to [bend left=10] (c);
    \draw[thick, ->, line width=0.75mm] (g1A1) to [bend left=10] (g1);
    \draw[thick, ->, line width=0.75mm] (yA1) to [bend left=10] (y);
    
    \draw[thick, line width=0.75mm] (uA) to [bend left=10] (uA2);
    \draw[thick, line width=0.75mm] (vA) to [bend left=10] (vA2);
    \draw[thick, line width=0.75mm] (wA) to [bend left=10] (wA2);
    \draw[thick, line width=0.75mm] (cA) to [bend left=10] (cA2);
    \draw[thick, line width=0.75mm] (g1A) to [bend left=10] (g1A2);
    \draw[thick, line width=0.75mm] (yA) to [bend left=10] (yA2);

    \node[shape=rectangle, dashed, line width=0.5mm, align=center, above left=2cm of w, text width=5cm] (box) {$\mathcal{G} = (\mathcal{U}, \mathcal{E}_S, \mathcal{E}_T)$};
    \node[circle, fill=white, below of=box, line width=0.5mm] (circ110) {};
    \node[circle, fill=white, below of=circ110, line width=0.5mm] (circ1) {};
    \node[circle, draw, fill=white, right= 0.75cm of circ1, line width=0.5mm] (circ2) {};
    \node[circle, draw, fill=white, left= 0.75cm of circ1, line width=0.5mm] (circ3) {};
    \draw[thick, solid, line width=0.5mm] (circ2) --  node [pos=0.5,above] {$e_S \in \mathcal{E}_S$} (circ3);
    \node [below of=circ1, text width=5cm, align=center] (circ111) {Spatial edge (object-to-object)};

    \node[circle, fill=white, below of=circ111, line width=0.5mm] (circ1111) {};
    \node[circle, fill=white, below of=circ1111, line width=0.5mm] (circ11) {};
    \node[circle, draw, fill=white, right= 0.75cm of circ11, line width=0.5mm] (circ21) {};
    \node[circle, draw, fill=white, left= 0.75cm of circ11, line width=0.5mm] (circ31) {};
    \draw[thick, ->, dashed, line width=0.5mm] (circ31) --  node [pos=0.5,above] {$\hat{e}_S \in \mathcal{E}_S$} (circ21);
    \node [below of=circ11, text width=5cm, align=center] (circ112) {Spatial edge (camera-to-object)};

    \node[circle, fill=white, below of=circ112, line width=0.5mm] (circ1121) {};
    \node[circle, fill=white, below of=circ1121, line width=0.5mm] (circ12) {};
    \node[circle, draw, fill=white, right= 0.75cm of circ12, line width=0.5mm] (circ22) {};
    \node[circle, draw, fill=white, left= 0.75cm of circ12, line width=0.5mm] (circ32) {};
    \draw[thick, ->, solid, line width=0.5mm] (circ32) --  node [pos=0.5,above] {$e_T \in \mathcal{E}_T$} (circ22);
    \node [below of=circ12, text width=5cm, align=center] (desc1) {Temporal edge};
    
    \draw[thick, dashed, line width=0.5mm] (box.north west)  -- (box.north east);
    \draw[thick, dashed, line width=0.5mm] (box.north west)  -- (desc1.south west);
    \draw[thick, dashed, line width=0.5mm] (desc1.south east)  -- (desc1.south west);
    \draw[thick, dashed, line width=0.5mm] (box.north east)  -- (desc1.south east);

  }
  
\end{tikzpicture}}
\vspace{-2.0mm}
\caption{An example of a spatio-temporal graph, CSTAG, capturing object-to-object and camera-to-object interactions over time.}
\label{fig:cstagFig}
\end{figure}

In an unrolled spatio-temporal graph (Figure \ref{fig:cstagFig}), the object nodes $u \in \mathcal{U}_o$ are connected with  undirected edges of the form $e_S = (u, v) \in \mathcal{E}_S$, while the camera node $u_c$ is connected to the object nodes with a directed edge $\hat{e}_S = (u_c, u) \in \mathcal{E}_S$. The nodes at adjacent time steps are connected through a directed temporal edge $e_T = (u_t, u_{t+1}) \in \mathcal{E}_T$. Each node $u \in \mathcal{U}_o$ has $L$ attributes, $A_{1}, \cdots, A_{L}$ and each attribute $A_l$ is associated with a feature vector $F_l$ with a dimension $d_l$.

\vspace{-4mm}
\paragraph{Node attributes:} Each node (except camera node $u_c$) in a graph corresponds to an object. Each node includes its \textit{local attributes} that represent the semantic characteristics of an object. These attributes are grouped into two different types: \textbf{Region-based attributes:} In a dynamic and unconstrained outdoor environment, objects can appear at random scales and at random locations. We measure region properties that capture the motion-independent statistics of the objects \cite{carreira2010constrained, lee2012discovering}. These properties include contour area, contour perimeter, pixel co-ordinates of the object in an image, centroid of the object, and bounding box dimensions. These features are independent of the object's appearance or motion.
 \textbf{Object-based attributes:} These properties include singular values (obtained using SVD) of an object's region, RGB histogram, HSV histogram, and the region corresponding to the object (in RGB color space as well as segmentation mask). These properties help in computing the gestalt properties \cite{carreira2010constrained} such as inter-object brightness similarity and inter-object energy similarity and are used only for computing the temporal edge.

\vspace{-4mm}
\paragraph{Edge attributes:} The edge attributes allow us to capture the interactions between the objects present in the image and the camera, both spatially and temporally. We can categorize the edges in a CSTAG into three categories: (1) object-to-object edges $e_S \in \mathcal{E}_S$, (2) camera-to-object edges $\hat{e}_S \in \mathcal{E}_S$, and (3) temporal edges $e_T \in \mathcal{E}_T$. Object-to-object edge features are measured between  the neighboring objects. These features include  pixel distance, counter-clockwise orientation with respect to the horizontal, and the amount of overlap with the bounding box of the neighboring object \cite{prabhu2015attribute}. Camera-to-object edge features are measured between each object in the image and the camera. These features include pixel distance, orientation, and shortest or projected distance with respect to the vertical. To link nodes with a temporal edge, we measure the change in the different selected node attributes; this is discussed next.

\vspace{-4mm}
\paragraph{Computing temporal edges:} The position of objects changes in an outdoor environment either due to their own movement or movement of the camera or both. Therefore, we need to map the graph nodes at time $t$ and $t+1$. Suppose that $u \in \mathcal{U}_t$ and $v \in \mathcal{U}_{t+1}$ are nodes  in graph $\mathcal{G}_{t}$ and $\mathcal{G}_{t+1}$ at times $t$ and $t+1$, respectively, that potentially represent the same object. The displacement of an object from time $t$ to $t+1$ is very small and therefore, these similar nodes must adhere to brightness and neighborhood consistency constraints \cite{gomila2003graph, nam2014online, mehta2016region}. If nodes $u$ and $v$ adhere to these constraints, then we connect these nodes with an edge $e_T = (u, v)$, otherwise,  $v$ becomes a new independent node in the CSTAG. We define these two constraints as follows:

\textit{Brightness consistency:} The change in appearance of an object from time $t$ to $t+1$ is very small. However, two or more nodes that are not close neighbors at time $t$ could be close neighbors at time $t+1$, say when two nodes are approaching each other. To make our method robust against such scenarios, we use a MAP-based template classification approach to identify similar nodes\footnote{To deal with occlusion, we follow contour-based occlusion detection method \cite{yilmaz2004contour, mehta2016region}.}. To do so, we optimize the following MAP estimate:
\begin{equation}
\resizebox{0.9\columnwidth}{!}{
$v^\star = \underset{u \in \mathcal{V}_t}{\argmax}\ \Pr(u \mid v), \ \text{where } \Pr(u \mid v) = \frac{\exp \left(f(v, u) \right)}{\underset{u' \in \mathcal{V}_t}{\sum} \exp \left(f(v, u')\right)}$
}
\label{eq:trackUp}
\end{equation}
where the function $\Pr$ is the probability of similarity between $u$ and $v$, and $\mathcal{V}_t$ denotes the set of all possible neighboring nodes of $v$ at time $t$. The function $f(v, u)$ is a function of three similarity measures: (1) color similarity \cite{yoo2009fast}, which is measured as a normalized correlation coefficient between the RGB attributes, (2) intensity similarity \cite{perez2002color}, which is measured as the Bhattacharya distance between the HSV histogram attributes, and (3) energy similarity \cite{liu2002svd}, which is measured as the L2-norm between the singular values of nodes $u$ and $v$. 

\textit{Neighborhood consistency:} The displacement of an object from time $t$ to $t+1$ is very small, and therefore, similar nodes must adhere to neighborhood consistency. Suppose that $v$ is the most similar node to $u$, determined using the brightness consistency constraints. If $\delta(v,u) < \tau$, where $\delta(v,u)$ is the change in the centroid of two nodes and $\tau$ is a predefined threshold, then the two nodes are similar. In our experiments, we found that the value of $\tau=20$ works best.
\vspace{-2.0mm}
\section{Identifying the Most Walkable Direction}
\label{ssec:costfunction}
\vspace{-2.0mm}
An outdoor environment is highly dynamic, and objects appear randomly at different times. It is challenging for visually impaired people to get around safely without colliding with objects. For example, it is difficult for blind people to change their paths if the sidewalk is blocked temporarily (e.g. due to  construction signs), thus they might hit the object blocking the sidewalk. To avoid such collisions, we identify an object (including ground) which is safe to follow using spatio-temporal object attributes. We predict the most walkable direction based on the safe-to-follow objects. We define a cost function that measures the safety of the direction and is modeled as a function of spatial and temporal attributes of the objects in the scene.
To learn the cost function, we minimize the following objective function:
\begin{equation}
u^\star  = \underset{u \in \mathcal{U}_o}{\argmin}\ C(u) 
\label{eq:edge}
\end{equation}
where $u^\star$ corresponds to the safe-to-follow object node.

Our cost function $C$ is a weighted sum of spatial cost $C_{sp}$ and temporal cost $C_{T}$ with the trade-off factor $\lambda$.
\begin{equation}
C = \lambda C_{sp} + (1 - \lambda ) C_{T}
\label{eq:cost}
\end{equation}
We compute the spatial cost $C_{sp}$ and the temporal cost $C_{T}$ for all the edges between the camera node $u_c$ and every object node $u \in  \mathcal{U}_o$ in the graph $\mathcal{G} = (\mathcal{U}, \mathcal{E}_S, \mathcal{E}_T)$:
\begin{equation}
\resizebox{0.9\columnwidth}{!}{
$C_{sp}(u, \mathbf{w}_{sp}) = \mathbf{w}_{sp} \cdot \mathbf{F}(u), \quad C_{T}(u, \mathbf{w}_{T}) = \mathbf{w}_T \cdot \delta \mathbf{F}(u)
$}
\label{eq:spCost}
\end{equation}
where $C_{sp}$ and $\mathbf{w}_{sp}$ are the spatial cost function and spatial weights, $C_{T}$ and $\mathbf{w}_T$ are the temporal cost function and temporal weights, $\mathbf{F}(u)$ is a feature vector that concatenate the features from node and edge attributes, and $\delta \mathbf{F}(u)$ is the change in the features of nodes $u$ over a time interval $k$.

\begin{figure}[t]
\centering
\begin{subfigure}[b]{0.45\columnwidth}
\centering
\includegraphics[width=\columnwidth]{samples1/frame_7070_overlay.jpg}
\vspace{-2.0mm}
\caption{\small{Label corresponding to Fig. \ref{fig:dataset}(e)}}
\end{subfigure}
\hfill
\begin{subfigure}[b]{0.5\columnwidth}
\centering
\includegraphics[width=\columnwidth]{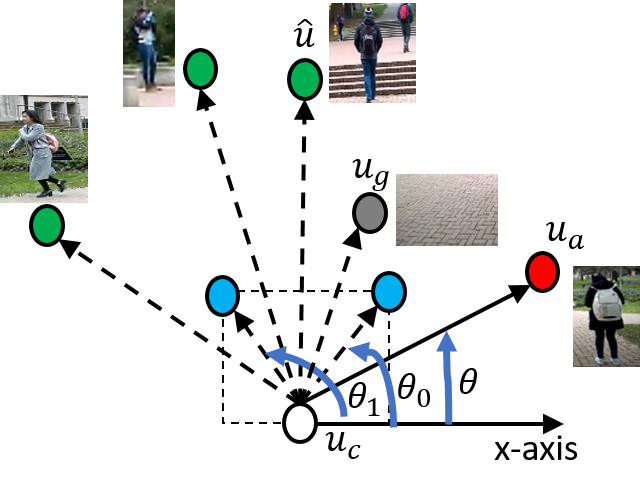}
\vspace{-2.0mm}
\caption{\small{Optimization scheme}}
\end{subfigure}
\setlength{\belowcaptionskip}{-4mm}
\caption{An example of the optimization scheme. The object corresponding to $u_a$ (predicted node) is close to the camera node $u_c$, so that a person holding the camera might collide with that object in the near future. The cost function must minimize the error between the predicted $u_a$ and the safest $\hat{u}$ node. The top-right and top-left corner of the safe walking area (blue region in (a)) make a angles $\theta_0$ and $\theta_1$ with the horizontal, while the predicted node $u_a$ makes an angle $\theta$ with the horizontal.}
\label{fig:reference}
\end{figure}

\textit{\textbf{Learning $\mathbf{w}_{sp}$, $\mathbf{w}_{T}$ and  $\lambda$:}} The objects in a video appear randomly, and there is a likelihood that there are no ambulatory objects in the frame for a certain duration. Furthermore, the information stored in a node is noisy due to noise in the segmentation model and distortions in the video. To make our model agnostic to such scenarios, we would like it to learn the spatial and temporal relationship between the node and edge attributes. Assume that we have the training data, $D_{train} = \{(\mathcal{G}_1, \theta_{01}, \theta_{11}), \cdots, (\mathcal{G}_M, \theta_{0M}, \theta_{1M}) \}$, that contains a graph $\mathcal{G}_t$ for each frame at time $t$, and the edges of the safe walking area denoted with the angles $\theta_{0t}$ and $\theta_{1t}$ (Figure \ref{fig:reference}).

Our system learns the spatial and temporal weights from the training data by minimizing the squared-error:
\setlength{\belowdisplayskip}{0pt} \setlength{\belowdisplayshortskip}{0pt}
\begin{equation}
\resizebox{0.9\columnwidth}{!}{
$\mathbf{w}_{sp}^\star = \underset{\mathbf{w}_{sp}}{\argmin} \frac{1}{M} \sum\limits_{i=1}^{M} \left(\mathbf{F}(\hat{u}_i) - \mathbf{w}_{sp} \cdot \mathbf{F}(u_i)\right)^2$
}
\end{equation}

\begin{equation}
\resizebox{0.9\columnwidth}{!}{
$\mathbf{w}_{T}^\star = \underset{\mathbf{w}_{T}}{\argmin} \frac{k}{M} \sum\limits_{\substack{1 \le i \le M,\\ k \mid i}} \left(\delta \mathbf{F}(\hat{u}_i) - \mathbf{w}_{T} \cdot \delta \mathbf{F}(u_i)\right)^2$
}
\end{equation}
where $\hat{u}$ is the node of a safe object randomly selected between $\theta_0$ and $\theta_1$ (see Figure \ref{fig:reference}). If there are no safe objects between $\theta_0$ and $\theta_1$,  the node corresponding to ground is selected.  The weight vectors $\mathbf{w}_{sp} \sim \mathcal{N}(0,\,1)$ and $\mathbf{w}_{T} \sim \mathcal{N}(0,\,1)$ are initialized randomly, and  the objective function is minimized using stochastic gradient descent (SGD). For learning the temporal weights, the values of $\delta \mathbf{F}$ between the $t^{th}$ frame and $(t+k)^{th}$ frame are measured. During training,  the value of $k$  is varied randomly between $5$ and $30$ so that the learned weights are  able to capture the short- and long-range temporal relations. 

Spatial and temporal weights are learned independently. The system first learns the spatial weights, $\mathbf{w}_{sp}$, and then uses them to learn the temporal weights, $\mathbf{w}_{T}$. We picked the trade-off factor, $\lambda$, empirically by minimizing the number of mistakes, $m$, where $m$ is 0 whenever the predicted $\theta \in [\theta_0, \theta_1]$ and 1 otherwise. In our experiments, $\lambda=0.3$ has the best performance.

\vspace{-2.0mm}
\section{Fast and Accurate Semantic Segmentation}
\label{sec:semanticSeg}
\begin{figure}[b!]
\setlength{\belowcaptionskip}{-4mm}
\centering
\resizebox{\columnwidth}{!}{\input{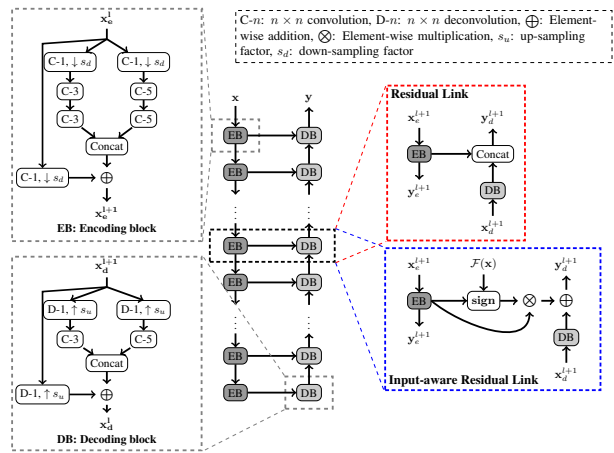}} 
\caption{Encoder-decoder network with the conventional residual \cite{ronneberger2015u, pinheiro2016learning} and the proposed input-aware residual links.}
\label{fig:architecture}
\end{figure}

\vspace{-2.0mm}
We use semantic segmentation to construct the spatio-temporal graph, which demands low-latency predictions. We extend the prior encoder-decoder networks for semantic segmentation \cite{shelhamer2017fully,badrinarayanan2017segnet,ronneberger2015u} and build a custom architecture that is both fast and accurate. Our network (called MSRSegNet) incorporates three new features: (1) multi-scale encoding blocks, (2) multi-scale decoding blocks, and (3) input-aware residual link for sharing the information between encoding and its corresponding decoding block. An overview of our network is given in Figure \ref{fig:architecture}.

\vspace{-4mm}
\paragraph{Multi-scale encoding blocks:} Following GoogLeNet \cite{szegedy2015going}, we use parallel feed-forward paths to aggregate features at different scales. Our network has symmetric paths where the depth and number of channels for each feed-forward path in a block are the same; these paths enable simultaneous  execution of multiple convolutional kernels at different scales, allowing aggregation of information with different effective receptive fields in the same time as using a single scale. Each path stacks convolutional layers to increase the effective receptive field \cite{simonyan2014very}. To improve the information flow inside the network, we add a skip-connection between input and output of the block \cite{he2016deep}. 

\vspace{-4mm}
\paragraph{Multi-scale decoding blocks:} The encoding network aggregates features at different spatial resolutions by performing convolution and down-sampling operations. The decoding network inverts the loss of resolution due to down-sampling operations, achieved using up-sampling operations (e.g. interpolation or deconvolution). Our decoding network stacks the multi-scale decoding blocks (see Figure \ref{fig:architecture}). Further, we use deconvolutional filters in our decoding blocks to learn the non-linear up-sampling operation. 

\begin{figure}[b!]
\setlength{\belowcaptionskip}{-4mm}
\centering
\setlength\tabcolsep{1pt}
\begin{tabular}{ccccc}
\small{RGB} & \small{Plain} & \small{Residual} & \small{IARC} &  \multirow{3}{*}{\includegraphics[width=0.02\columnwidth, valign=b]{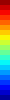}}\\
\small{Image} & \small{Enc-Dec} & \small{Enc-Dec} & \small{Enc-Dec} & \\
\includegraphics[width=0.22\columnwidth, valign=c]{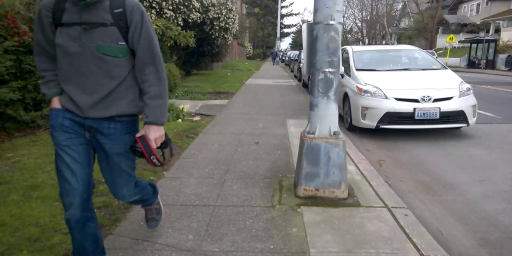} &  
\includegraphics[width=0.22\columnwidth, valign=c]{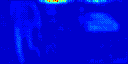} &
\includegraphics[width=0.22\columnwidth, valign=c]{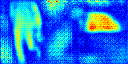} &
\includegraphics[width=0.22\columnwidth, valign=c]{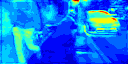} & 
\end{tabular}
\caption{Visualization of decoded feature maps. Learned filters with input-aware residual connections (IARC) help to reveal the input-relevant features (e.g. boundaries of a pole) which are not revealed in other two networks. Red and blue colors denote the maximum and minimum values in the feature map respectively.}
\label{fig:visual}
\end{figure}

\vspace{-4mm}
\paragraph{Input-aware residual connection (IARC):} CNN feature maps relate to power spectral density, where higher-frequency components decay with the depth of the network, leaving mainly the low-frequency components at lower spatial resolutions. Simply up-sampling these low-frequency components generates coarse results that can be improved by combining the feature maps from different levels of the network to include details about both low- and high-frequency components \cite{shelhamer2017fully, ronneberger2015u, pinheiro2016learning, ghiasi2016laplacian}. Though fusing feature maps from different levels helps to refine the segmentation masks, they still struggle at recovering some of the fine-grained information which is lost due to the down-sampling operations (Figure \ref{fig:visual}). To tackle this, we introduce an input-aware residual connection that reinforces the input at different spatial levels, allowing us to learn \textit{input-relevant features} (e.g. object boundaries), which in turn refines the coarse feature maps for better predictions. Our input-aware residual connection first identifies the relevant high-frequency components between an input image and the encoded feature map (using a sign function) and then suppresses the low-frequency components from the encoded feature map using multiplicative gating. The resultant map is then used to refine the decoded feature map\footnote{Closely related architecture was used in \cite{ghiasi2016laplacian}, where the feature maps at different levels were fused using a masking operation. Such an architecture involves stage-wise training, while our method eliminates it. Our method is $\approx 20 \times$ faster than \cite{ghiasi2016laplacian} while delivering similar performance.}. IARC can be mathematically defined as:
\begin{equation}
 \mathbf{y}^{l+1}_d = \Big(\mathbf{sign}\left(\mathbf{y}_e^{l+1}\ ,\  \mathcal{F}(\mathbf{x})\right) \otimes \mathbf{y}_e^{l+1}\Big) \oplus \mathbf{y}_d^{l+1}
\end{equation} 
where $\mathcal{F}(\mathbf{x})$ represents the input-aware feature mapping. $\mathcal{F}(\mathbf{x})$ is a composite function consisting of a $3\times 3$ average pooling operation that sub-samples the $\mathbf{x}$ to the same spatial dimensionality as $\mathbf{y}_e^{l+1}$, followed by a $1\times 1$ and $3\times 3$ convolution, where $1\times 1$ convolution projects the resultant sub-sampled map to the same dimensionality (channel-wise) as of $\mathbf{y}_e^{l+1}$. The $\otimes$ operation and $\oplus$ denotes the element-wise multiplication and addition operations respectively, and $\mathbf{sign}$ is an indicator function defined as: $\mathbf{sign}(a\ ,\ b) = \mathds{1}(sgn(a) == sgn(b))$.

\vspace{-2.0mm}
\section{Experiments}
\vspace{-2.0mm}
In this section, we describe our experimental set-up and compare the performance of the proposed method with (1) different baselines and (2) different segmentation methods. We also analyze the performance of MSRSegNet on the publicly available segmentation datasets.
\vspace{-2.0mm}
\subsection{Experimental set-up}
\vspace{-2.0mm}
\paragraph{Dataset details:} We used two different hand-held mobile devices (Windows Lumia 960 and iPhone 6s) to capture the videos at a resolution of 720p. We collected videos over an overall area of 1.8 miles (Figure \ref{fig:areamap}), a walking distance of about 40 minutes for a normal sighted person. The videos were collected at random times with random duration (2 to 10 minutes) on different days and under different weather conditions, which ensured that our dataset has sufficient variety in terms of background, objects, and object activities. These videos were collected in unconstrained outdoor environments and are challenging because: (1) the objects appearing in these videos are different (e.g. in size, shape, scale, color, speed, and occlusion), (2) the camera view-point and illumination changes, (3) motion blur, (4) different ground types, and (5) traffic conditions. The total duration of collected videos was about 70 minutes. After removing video snippets that were similar in appearance or had little or no activity, a total of 40,236 frames remained and were segmented into smaller video snippets, each having about 200 frames at 20 fps. The resulting  201 video snippets were then grouped into three categories: (1) easy:  videos with 0-4 ambulatory objects, (2) moderate:  videos with 4-8 ambulatory objects, and (3) hard: videos with more than 8 ambulatory objects.

\vspace{-4mm}
\paragraph{Annotation details:} To train a model that could identify the most walkable direction while avoiding collisions with objects, we need ground truth data  containing information about the safety level of objects. Some large publicly annotated datasets (e.g. MS-COCO \cite{lin2014microsoft} and PASCAL VOC \cite{everingham2015pascal}) are used for object localization (e.g. person and vehicle), but lack information about the safety level of objects, which is crucial for training a model for safe navigation in an unconstrained outdoor environment. With this goal, we designed an annotation label set and asked human annotators to watch a video for a few seconds and annotate the objects with the following safety levels: (1) safe: object is safe to follow, (2) avoid (high, medium, low): object should be avoided, and (3) best area on the ground; safe for the next step. The best ground area markings are capped to about 120 pixels along the y-axis from the camera node, which translates to about 5 meters in real-world (measured using camera calibration) and is considered a safe distance between the object and the blind person \cite{pradeep2010robot}. Sample annotations are shown in Figure \ref{fig:dataset}. 

\begin{figure}[b!]
\setlength{\belowcaptionskip}{-4mm}
\centering
\begin{subfigure}[b]{0.49\columnwidth}
\includegraphics[height=70px]{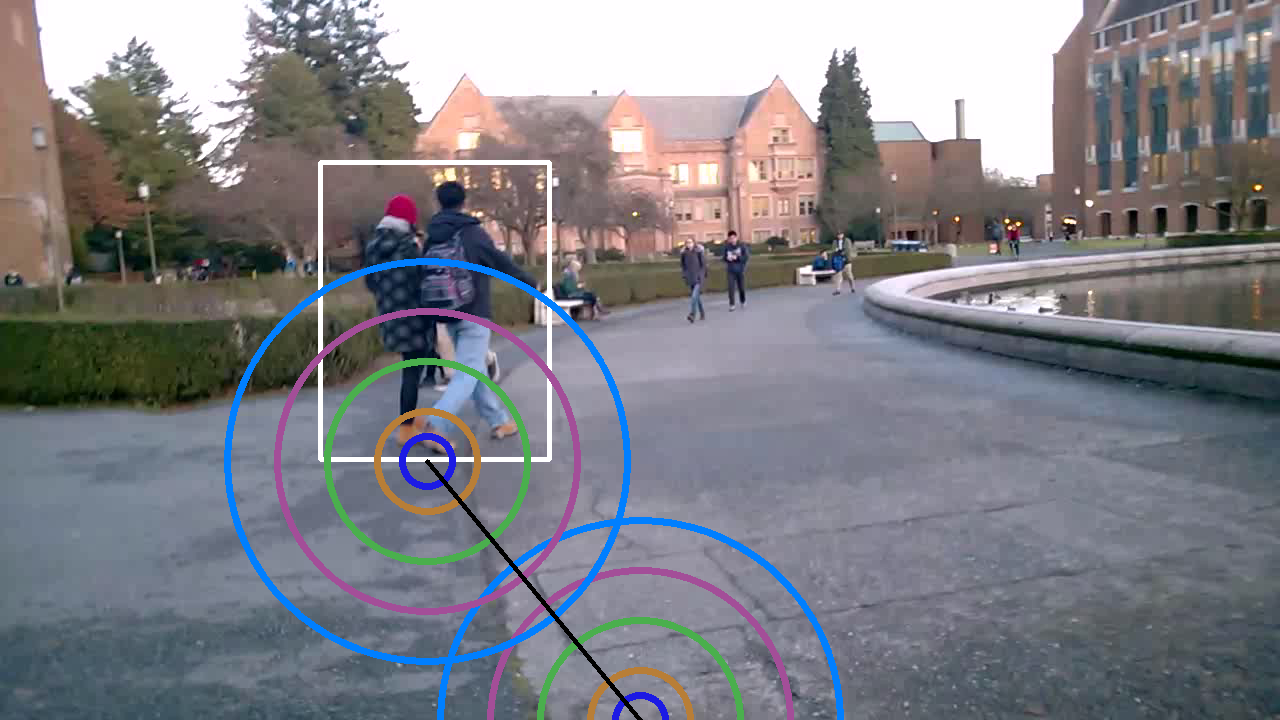} 
\caption{}
\label{fig:radii}
\end{subfigure}
\begin{subfigure}[b]{0.49\columnwidth}
\includegraphics[height=70px]{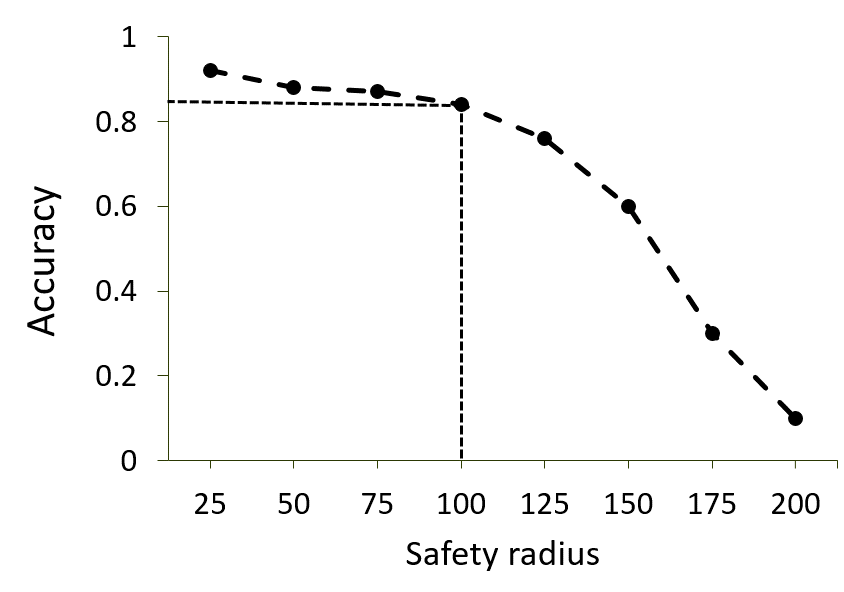} 
\caption{}
\label{fig:rVsAcc}
\end{subfigure}
\caption{Regions corresponding to object and camera node at different safety radii $r = \{\textcolor{c1}{25}, \textcolor{c2}{50}, \textcolor{c3}{100}, \textcolor{c4}{150}, \textcolor{c5}{200}\}$ in pixels (a) along with their impact on accuracy (b) are shown.}
\end{figure}

\vspace{-4mm}
\paragraph{Evaluation metric:} The accuracy metric should account for both safety as well as collision, which are defined as: 
\textit{Safety:} Each video frame in our dataset has a category that is marked as ``safe ground area". We first measure whether the predicted direction is safe or not. To do so, we measure the angle $\theta$ between the predicted node $u^\star$ and the horizontal axis, since the camera node is located on that axis. If $\theta \in \left[ \theta_0, \theta_1 \right]$, then the predicted object node is safe (Figure \ref{fig:reference}). 
 \textit{Collision avoidance:} We measure the distance between the predicted node $u^\star$ from the camera node $u_c$. If $R_C \cap R_O = \phi$, then we say that the object is at sufficient distance from the camera and chances to collide in the next time step are negligible. Here, $R_C$ and $R_O$ denotes the safe regions corresponding to $u_c$ and $u^\star$, each with a safety radius of $r$ (see Figure \ref{fig:radii}).

We then define the discrete accuracy metric as:
\begin{equation}
\resizebox{0.8\columnwidth}{!}{
$\text{Accuracy}  = \begin{cases}
1, & \text{if } \theta \in [\theta_0, \theta_1] \text{ and } R_C \cap R_O = \phi \\
0, & \text{otherwise}
\end{cases}$
}
\label{eq:acc}
\end{equation}

\fboxsep=0.00mm
\fboxrule=0.25mm
\begin{figure*}[t!]
\centering
\begin{subfigure}[b]{0.5\columnwidth}
\fcolorbox{green}{green}{\includegraphics[width=0.9\columnwidth]{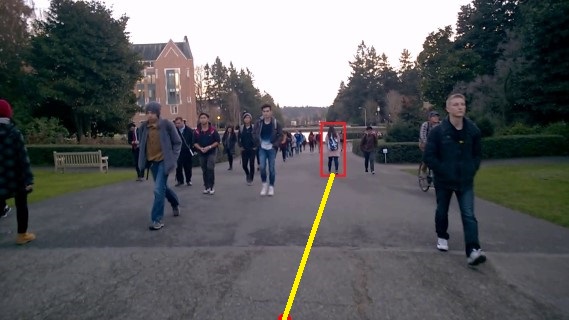}} 
\caption{}
\end{subfigure}
\begin{subfigure}[b]{0.5\columnwidth}
\fcolorbox{green}{green}{\includegraphics[width=0.9\columnwidth]{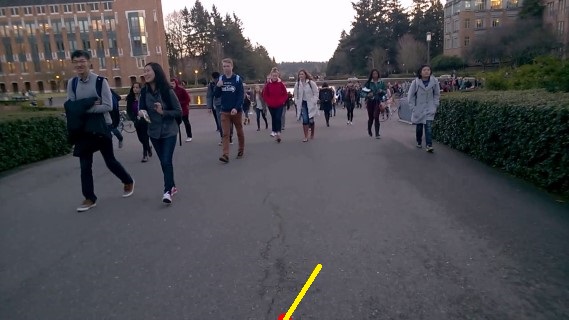}}
\caption{}
\end{subfigure}
\begin{subfigure}[b]{0.5\columnwidth}
\fcolorbox{green}{green}{\includegraphics[width=0.9\columnwidth]{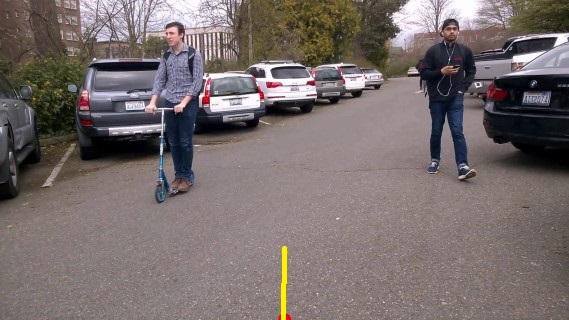}}
\caption{}
\end{subfigure}
\begin{subfigure}[b]{0.5\columnwidth}
\fcolorbox{green}{green}{\includegraphics[width=0.9\columnwidth]{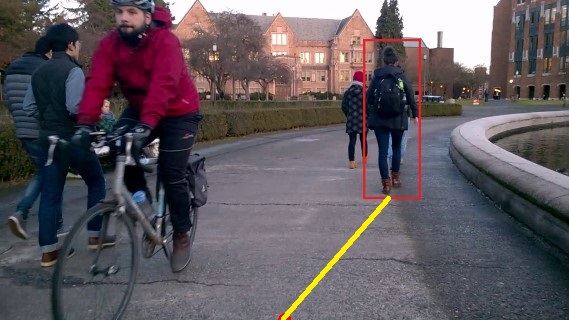}} 
\caption{}
\label{fig:corrPred}
\end{subfigure}
\begin{subfigure}[b]{0.5\columnwidth}
\fcolorbox{green}{green}{\includegraphics[width=0.9\columnwidth]{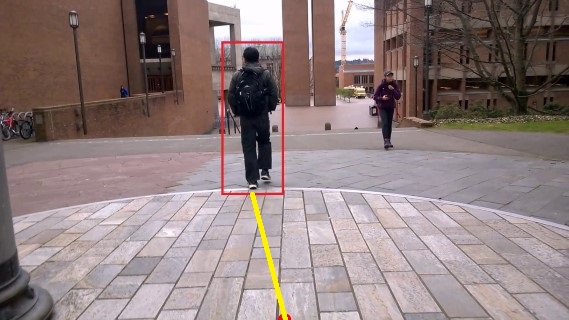}}
\caption{}
\end{subfigure}
\begin{subfigure}[b]{0.5\columnwidth}
\fcolorbox{green}{green}{\includegraphics[width=0.9\columnwidth]{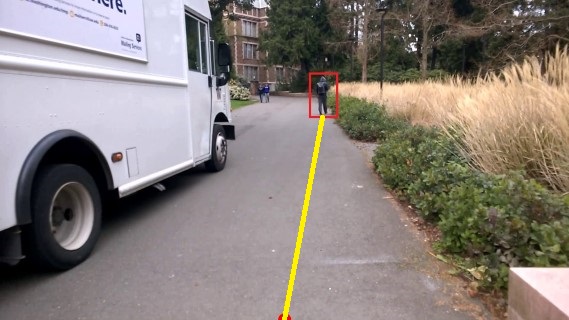}} 
\caption{}
\end{subfigure}
\begin{subfigure}[b]{0.5\columnwidth}
\fcolorbox{green}{green}{\includegraphics[width=0.9\columnwidth]{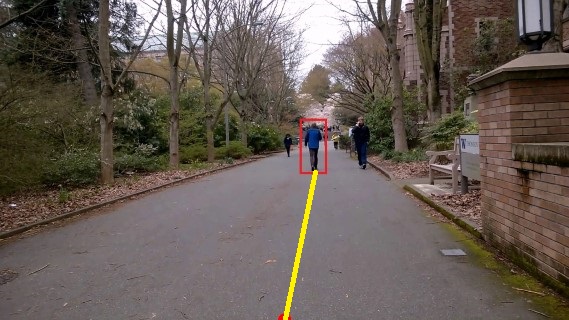}} 
\caption{}
\end{subfigure}
\begin{subfigure}[b]{0.5\columnwidth}
\fcolorbox{green}{green}{\includegraphics[width=0.9\columnwidth]{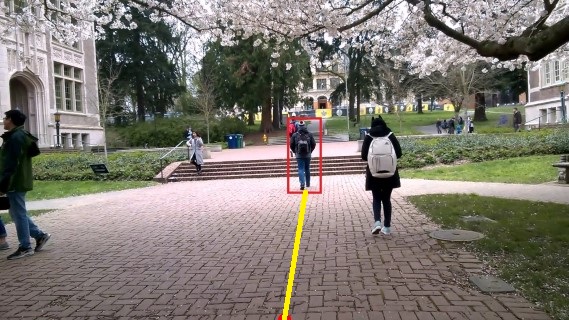}}
\caption{}
\end{subfigure}
\begin{subfigure}[b]{0.5\columnwidth}
\fcolorbox{green}{green}{\includegraphics[width=0.9\columnwidth]{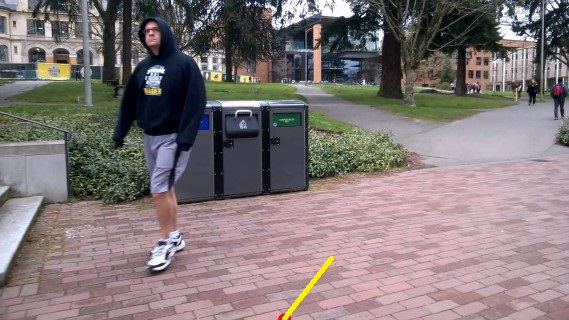}} 
\caption{}
\end{subfigure}
\begin{subfigure}[b]{0.5\columnwidth}
\fcolorbox{green}{green}{\includegraphics[width=0.9\columnwidth]{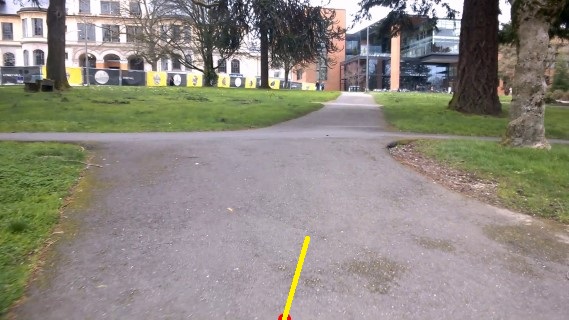}} 
\caption{}
\end{subfigure}
\begin{subfigure}[b]{0.5\columnwidth}
\fcolorbox{green}{green}{\includegraphics[width=0.9\columnwidth]{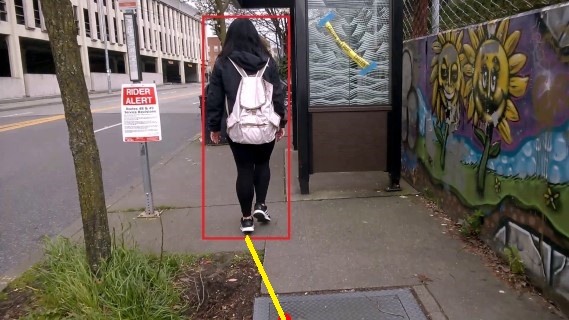}} 
\caption{}
\end{subfigure}
\begin{subfigure}[b]{0.5\columnwidth}
\fcolorbox{green}{green}{\includegraphics[width=0.9\columnwidth]{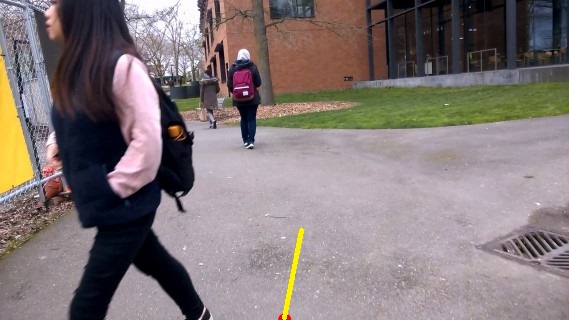}} 
\caption{}
\end{subfigure}
\begin{subfigure}[b]{0.5\columnwidth}
\fcolorbox{green}{green}{\includegraphics[width=0.9\columnwidth]{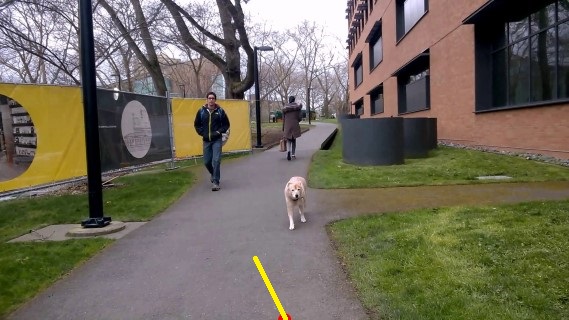}}
\caption{}
\end{subfigure}
\begin{subfigure}[b]{0.5\columnwidth}
\fcolorbox{red}{red}{\includegraphics[width=0.9\columnwidth]{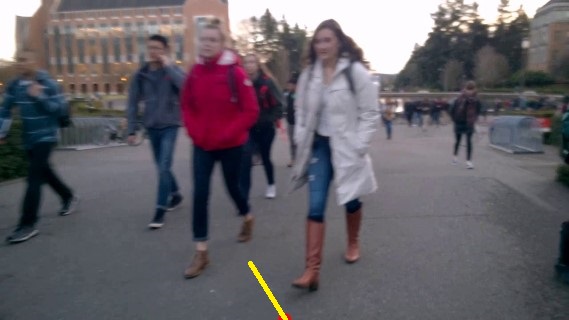}} 
\caption{}
\end{subfigure}
\begin{subfigure}[b]{0.5\columnwidth}
\fcolorbox{red}{red}{\includegraphics[width=0.9\columnwidth]{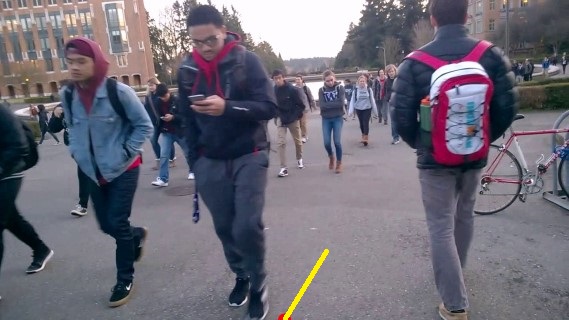}}
\caption{}
\end{subfigure}
\begin{subfigure}[b]{0.5\columnwidth}
\fcolorbox{red}{red}{\includegraphics[width=0.9\columnwidth]{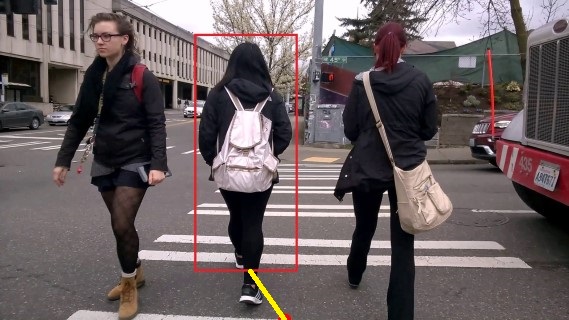}} 
\caption{}
\label{fig:falsePred}
\end{subfigure}
\setlength{\belowcaptionskip}{-4mm}
\caption{Qualitative results suggest that our method adapts to the dynamics of the environment (e.g. street view, parking lot, and different objects including ground surfaces) and is able to identify the most walkable direction. The yellow line indicates the most walkable direction predicted by our method. The images with green and red borders correspond to the positive and negative predictions, respectively. For negative predictions in this figure, our algorithm predicted the object correctly (ground or ambulatory), but violated the collision constraint. For visualization, we have shown only the bounding box of the object that is safe to follow. }
\label{fig:exampleResults}
\end{figure*}

\vspace{-4mm}
\paragraph{Training details:} We split our dataset randomly into training (50 videos) and test sets (about 150 videos). We used the training set to learn the cost function $C$ (Eq. \ref{eq:cost}) using 5-fold cross validation. We used a machine with Intel CPU i7-6700K, 16 GB RAM, and NVIDIA TitanX GPU. 

\vspace{-4mm}
\paragraph{Baselines:} We design different baselines to study the effect of graph representation and the cost function: (1) \textit{random:} a node is selected randomly in the graph corresponding to an object, (2) \textit{max:} a node with a maximum distance to the camera is selected, (3) \textit{random with temporal:} a random node is selected among the nodes whose direction is the same as the camera wearer, (4) \textit{max with temporal:} same as (3), but the node with maximum distance to the camera node is selected, and (5) \textit{non-graph:} In order to study the effect of our graph-based representation that is build on semantic segmentation, we design a baseline that takes RGB image as an input and predicts the direction corresponding to the safe-to-follow object. We fine-tuned a CNN and LSTM-based method \cite{xu2015show} to predict an angle $\theta$ between the safe object and the horizontal. For this baseline, our training data consists of an image and a sequence of angles $\theta \in \left[\theta_0, \theta_1\right]$ between the safe object node and the horizontal for the current and next 9 frames.

\vspace{-4mm}
\paragraph{Implementation details:} We used tracking to extract temporal information from the video instead of segmenting every video frame, employing the GOTURN \cite{held2016learning} tracking method, because it provided the highest tracking accuracy over competitors.
To find the safety radius for our evaluation, we varied the value of $r$ from 25 to 200 pixels. The impact of safety radius on the accuracy is shown in Figure \ref{fig:rVsAcc}. The accuracy of our method dropped as we increased the safety radius. Our method achieved an accuracy of about 84\% at a safety radius of 100 pixels. In our camera calibration experiments, we found that a radius of 100 pixels around the camera node correspond to an actual distance of about 4.5 meters along the principal axis from the person holding the camera, which is considered a safe distance between the camera and the object \cite{pradeep2010robot}.
\vspace{-2.0mm}
\subsection{Comparison with different baselines}
\vspace{-2.0mm}
Table \ref{tab:costEval} shows the accuracy of our method against the baselines. We can see that the learned cost function plays an important role in predicting the most walkable direction. The accuracy of the system improved by about 4\% when temporal data is incorporated in the cost function. Also, the CSTAG allows reasoning about the semantically rich contextual information from the scene, which is difficult with non-graph-based approaches. Therefore, it has higher accuracy in comparison to the non-graph-based baseline. Figure \ref{fig:exampleResults} shows qualitative examples of success and failure cases of our method. For example, in Figure \ref{fig:corrPred}, our method correctly identified the safe-to-follow object and not the bicyclists approaching the camera. However, in some cases (e.g. Figure \ref{fig:falsePred}), our method correctly predicted the safest node but violated the collision constraint since the identified object was very close to the camera.

\begin{table}[b!]
\setlength{\belowcaptionskip}{-4mm}
\centering
\resizebox{0.9\columnwidth}{!}{
\begin{tabular}{l|c|c|c|c}
\toprule
\multirow{2}{*}{\textbf{Edge Selection Criteria}}   & \multicolumn{3}{|c|}{\textbf{Accuracy per Video Category}} & \textbf{Overall} \\
\cmidrule{2-4}  & \textbf{Easy} & \textbf{Moderate} & \textbf{Hard} & \textbf{Accuracy} \\
\midrule
Non-graph & 0.82 & 0.41 & 0.24 & 0.49\\
\midrule
Random       & 0.87   & 0.38   & 0.20 & 0.48    \\
Random (w/ temporal)  & 0.91   & 0.49   & 0.27 & 0.56 \\
\midrule
Max. distance        & 0.90   & 0.42   & 0.28 & 0.53  \\
Max. distance (w/ temporal)    & 0.94   & 0.51   & 0.36 & 0.60 \\
\midrule
Min. spatial cost (Eq. \ref{eq:spCost})   & 0.93 & 0.87 & 0.60 & 0.80    \\
Min. spatio-temporal cost (Eq. \ref{eq:cost})  & \textbf{0.95}   & \textbf{0.88}   & \textbf{0.68} & \textbf{0.84}    \\
\bottomrule
\end{tabular}
}
\caption{This table compares the performance of the proposed method with different baselines. Our new method that uses minimum spatio-temporal cost achieved the best results.}
\label{tab:costEval}
\end{table}

\vspace{-2.0mm}
\subsection{Impact of segmentation methods} 
\vspace{-2.0mm}
Table \ref{tab:segMethodCompare} compares the performance of three segmentation methods on our task: (1) FCN-8s \cite{shelhamer2017fully}, (2) RefineNet \cite{lin2016refinenet}, and (3) MSRSegNet (Section \ref{sec:semanticSeg}) that were trained on the PASCAL Context dataset \cite{mottaghi_cvpr14}\footnote{ We choose PASCAL Context dataset because it had a good mix of the indoor and outdoor classes, which were hard to find in other datasets such as CamVid and Cityscapes, for example, stairs. }. We choose FCN-8s and RefineNet, because FCN-8s was fast but less accurate, while RefineNet was accurate but less fast. FCN-8s outperformed the simple baselines by a substantial margin. With RefineNet and MSRSegNet as segmentation methods, we attained similar accuracy. Though the segmentation accuracy of RefineNet on the PASCAL Context dataset is about 2\% higher than MSRSegNet, the impact on the accuracy on our dataset was negligible. This was likely because RefineNet performed well on the small objects (such as bottles), which were not significantly important for our task. Note that MSRSegNet is more than $21\times$ faster than RefineNet while delivering competitive accuracy.
\vspace{-2.0mm}
\begin{table}[t!]
\setlength{\belowcaptionskip}{-4mm}
\centering
\resizebox{\columnwidth}{!}{
\begin{tabular}{l|c|c|c|c}
\toprule
\textbf{Segmentation}  & \multicolumn{3}{|c|}{\textbf{Accuracy per Video Category}} & \textbf{Overall}   \\
\cmidrule{2-4}
\textbf{Method}  & \textbf{Easy} & \textbf{Moderate} & \textbf{Hard} & \textbf{Accuracy}  \\
\midrule
FCN-8s \cite{shelhamer2017fully} & 0.92   & 0.78   & 0.59 & 0.76   \\
\midrule
RefineNet \cite{lin2016refinenet} & \textbf{0.95}   & \textbf{0.89}   & \textbf{0.68} & \textbf{0.84}   \\
\midrule
MSRSegNet (Section \ref{sec:semanticSeg}) & \textbf{0.95}   & 0.88   & \textbf{0.68} & \textbf{0.84}  \\
\bottomrule
\end{tabular}
}
\caption{We compare the performance of different segmentation methods on the minimum spatio-temporal cost (Eq. \ref{eq:cost}) used to identify most walkable object.}
\label{tab:segMethodCompare}
\end{table}

\vspace{-2mm}
\paragraph{Results on the PASCAL Context dataset:}
We trained our segmentation network using SGD with an initial learning rate of 0.01 and decay of 10 every 30 epochs. We used a weight decay of 0.0005 and momentum of 0.9. To reduce the internal-covariate-shift problem~\cite{ioffe2015batch:eke}, we applied batch normalization (BN) \cite{ioffe2015batch:eke} after every convolutional and deconvolutional layer. We used spatial dropout \cite{tompson2015efficient:eke}, randomized ReLU (RReLU) \cite{xu2015empirical:eke}, and augmentation (scaling, cropping, and flipping) to prevent over-fitting. We measured the accuracy as mean region intersection over union (mIOU).

The PASCAL Context dataset contains 60 classes (including background) and provides whole scene segmentation of the PASCAL VOC images. The proposed method  (see Table \ref{tab:pascon}) is faster than the state-of-the-art methods while delivering competitive accuracy. To further show the effectiveness of the proposed method, we also trained our system on a widely used segmentation framework: the PASCAL VOC dataset. Our method attained a mIOU of 81.01 (see \cite{pasAnnony2}) while running at 21 frames per second. In particular, our method can run at different image resolutions, while providing an easy trade-off between speed and accuracy. At an image resolution of $224\times 224$, our method attains a mIOU of 67.12 (which is the same as the FCN-8s), but runs at a speed of 60 fps (see \cite{pasAnnony1}). For more detailed studies, please see Section \ref{sec:pascalSec}.

\begin{table}[t!]
\centering
\resizebox{0.8\columnwidth}{!}{
\begin{tabular}{l|cc|c|c}
\toprule
\textbf{Segmentation} & \multicolumn{2}{c|}{\textbf{Additional aids}} & \multirow{2}{*}{\textbf{mIOU}} & \textbf{Speed} \\
\cline{2-3}
\textbf{Framework} & \textbf{COCO} & \textbf{CRF} & & (in fps) \\
\midrule
FCN-8s \cite{shelhamer2017fully} & & & 37.8 & 10 \\
CRFasRNN \cite{crfasrnn_ICCV2015} & & \checkmark & 39.3 & $<$ 1 \\
DeepLab-v2 \cite{chen2016deeplab} & \checkmark & \checkmark & 45.7 & $<$ 1 \\
RefineNet \cite{lin2016refinenet} &  &  & \textbf{47.1} & $<1$ \\
MSRSegNet (Ours) & &  & 44.7 & 21 \\
\bottomrule
\end{tabular}
}
\setlength{\belowcaptionskip}{-3mm}
\caption{Segmentation frameworks on the PASCAL Context dataset. Our MSRSegNet is more than 21 times faster than RefineNet and almost as accurate, while the next fastest method (FCN-8s) is much less accurate. Inference speed was measured for an input of $512 \times 512$ on NVIDIA TitanX GPU.}
\label{tab:pascon}
\end{table}
\vspace{-4mm}
\paragraph{Speed:} Existing networks use VGG-16 and ResNet-101 as base feature extractors. These feature extractors are either wide or deep and therefore, they are slow (Table \ref{tab:compArchBase}). We exploited the recent advancements in hardware technology (e.g. TitanX can execute tera FLOPS) to make our network fast. Our network executes two convolutional kernels simultaneously in the same time as opposed to a single convolutional kernel in VGG-16 and ResNet-101. A custom and efficient base feature extractor along with a light-weight decoder makes our network fast. Furthermore, in contrast to computationally expensive post-processing methods such as CRF, we used a novel residual connection IARC that improved the accuracy without drastically reducing inference speed. In our experiments, we found that IARC improved the accuracy of a plain encoder-decoder network by about $4\%$ across different base feature extractors, which was $2\%$ more than the residual connections.

\begin{table}[t!]
\centering
\resizebox{\columnwidth}{!}{
\begin{tabular}{l|c|c|c|c|c}
\toprule
 & \textbf{FLOPS}$\dagger$ & \textbf{Parameters}$\dagger$ & \textbf{Memory}$\dagger$ & \textbf{Inference Time} & \textbf{top-5 accuracy} \\
  & (in billion) & (in million) & (in MB) & (in ms) & (in \%) \\
 \midrule
VGG-16 & 15.44 & 117.43 & 35.19 & 33.37& 90.67\\
ResNet-101 & \textbf{7.57} & \textbf{42.39} & 73.36& 80.38 & \textbf{93.95}\\
Ours & 13.02 & 100.58 & \textbf{18.06} & \textbf{22.11} & 91.53\\
\bottomrule
\end{tabular}
}
\setlength{\belowcaptionskip}{-5mm}
\caption{Comparison between different base feature extractors for an input image of size $224\times 224$ on the ImageNet validation set. Our network has a depth of 18 and is almost $1.5\times$ and $4\times$ faster than VGG-16 and ResNet-101, respectively, while delivering competitive accuracy. $\dagger$ The fully connected layers are not considered. For more details, see Table \ref{tab:compVRO}.}
\label{tab:compArchBase}
\end{table}

\vspace{-2.0mm}
\section{Conclusion}
\vspace{-2.0mm}
In this paper, we propose an approach for identifying the most walkable direction for navigation using a hand-held camera in an unconstrained outdoor environment. We introduce a new dataset consisting of approximately 40,000+ annotated frames. Our system achieves an accuracy of 84\% when tested with this dataset, while running at about 21 fps on TitanX device and 5 fps on a low-power device. Our system is semantic-aware; therefore, it can be used by the visually impaired for target-specific querying such as locating and navigating to a nearby trash can. Furthermore, our system can be easily integrated with GPS/SLAM-based localization methods to provide source-to-destination navigation cues with collision avoidance. 

\vspace{-2.0mm}
\section{Acknowledgement}
\vspace{-2.0mm}
We gratefully acknowledge the support of NVIDIA Corporation with the donation of the Titan X Pascal GPU used for this research. We also thank Dr. Ezgi Mercan and Dr. Anat Caspi for their thorough and helpful comments.

\clearpage

\appendix

\section{Results on the PASCAL VOC 2012 Dataset}
\label{sec:pascalSec}
The PASCAL VOC 2012 \cite{everingham2015pascal} is a well-known segmentation dataset and contains annotations for 21 classes including background. In this section, we first compare the performance of the proposed input-aware residual links with different base feature extractors. We then compare the proposed method with the state-of-the-art methods.

\vspace{-4mm}
\paragraph{Impact of input-aware residual connections:} Table \ref{fig:iasc} compares the performance of three encoder-decoder architectures with different base feature extractors: (1) plain encoder-decoder network, (2) plain network with residual links, and (3) plain network with input-aware residual links. We can see that IARC improved the accuracy of a plain encoder-decoder network by about $4\%$ across different base feature extractors, which was $2\%$ more than the residual connections. Note that ResNet-50 was about 2\% more accurate than our method, but was $3\times$ slower.

\begin{table}[b!]
\centering
\resizebox{0.8\columnwidth}{!}{
\begin{tabular}{ccc|c|c}
\toprule
 \multirow{2}{*}{\textbf{Plain}} & \textbf{Residual} & \textbf{Input-aware} & \multirow{2}{*}{\textbf{mIOU}}& \textbf{Speed} \\
   & \textbf{link} & \textbf{residual link} & & (in fps) \\
\midrule
\multicolumn{5}{c}{VGG-16 \cite{simonyan2014very} as the base model} \\
 \checkmark & & & 57.7 & 34 \\
 \checkmark & \checkmark & & 59.1  & 33  \\
 \checkmark &  & \checkmark & 61.3 & 25  \\
\midrule
\multicolumn{5}{c}{ResNet-50 \cite{he2016deep} as the base model}\\
 \checkmark & & & 62.1  & 31 \\
 \checkmark & \checkmark & & 64.7 & 28  \\
 \checkmark &  & \checkmark & \textbf{67.2} & 20 \\
\midrule
\multicolumn{5}{c}{Our classification network as the base model}\\
\checkmark & & & 60.7  & 75 \\
\checkmark & \checkmark & & 61.9 & 72  \\
\checkmark &  & \checkmark & 65.23 & \textbf{60} \\
\bottomrule
\end{tabular}
}
\caption{Performance of different encoder-decoder networks on PASCAL VOC 2012 validation set. All networks were trained with an input image size of $224 \times 224$. Inference speed is measured on NVIDIA PASCAL TitanX GPU and averaged over 100 trials on an input image having dimension $224 \times 224$.}
\label{fig:iasc}
\end{table}

\begin{table}[b!]
\resizebox{\columnwidth}{!}{
\begin{tabular}{l|ccc|cc}
\toprule
\textbf{Method} & \textbf{COCO} & \textbf{Object Proposals} & \textbf{CRF} & \textbf{mIOU} & \textbf{Speed} \\
\midrule
\multicolumn{6}{c}{VGG-16 as the base model}\\
SegNet \cite{badrinarayanan2017segnet} & & &  & 59.1 & 10 \\
FCN-8s \cite{shelhamer2017fully} & & & & 67.2 & 10 \\
CRFasRNN \cite{crfasrnn_ICCV2015} & \checkmark & & \checkmark & 74.7 & $<$ 1 \\
DeConvNet \cite{noh2015learning} & & \checkmark &   & 69.6 & $<$ 1 \\
DeConvNet \cite{noh2015learning} & & \checkmark & \checkmark  & 72.5 & $<$ 1 \\
Dilation-8 \cite{yu2015multi} & \checkmark & &  & 73.5 & $<$ 1 \\
Dilation-8 \cite{yu2015multi} & \checkmark & & \checkmark & 73.5 & $<$ 1 \\
\midrule
\multicolumn{6}{c}{ResNet-101 as the base model} \\
DeepLab-v2 \cite{chen2016deeplab} (val) & \checkmark & & & 75.4 & 5 \\
DeepLab-v2 \cite{chen2016deeplab} & \checkmark & & \checkmark & 79.7 & $<$ 1 \\
PSPNet \cite{zhao2016pyramid} & \checkmark & &  & \textbf{85.4} & $<$ 1 \\
RefineNet \cite{lin2016refinenet} & \checkmark & &  & 82.4 & $<1$ \\
LRR \cite{ghiasi2016laplacian} & \checkmark & &  & 78.7 & $<1$ \\
\midrule
\multicolumn{6}{c}{Our custom classification network as the base model} \\
Ours $224 \times 224$ $\dagger$ & \checkmark & &  & 67.12 & \textbf{60} \\
Ours $320 \times 320$ (val) & \checkmark & &  & 68.87 & 43 \\
Ours $448 \times 448$ (val) & \checkmark & &  & 71.32 & 27 \\
Ours $512 \times 512$ $\ddagger$ & \checkmark & &  & 81.01 & 21 \\
\bottomrule
\end{tabular}
}
\caption{Segmentation frameworks on the PASCAL VOC 2012 test dataset. The proposed method is faster than previous work, while delivering competitive accuracy. The proposed method can run at different image resolutions for an easy trade-off between accuracy and speed. Inference speed is measured on NVIDIA TitanX GPU. Models trained using MatConvNet were first ported into Caffe and then we measured the inference speed. Result links to the VOC evaluation server: $\dagger$ \url{http://host.robots.ox.ac.uk:8080/anonymous/IJV89W.html} $\ddagger$ \url{http://host.robots.ox.ac.uk:8080/anonymous/NQRTFB.html}}
\label{tab:pascal}
\end{table}
\begin{figure}[b!]
\centering
\begin{subfigure}[b]{0.2\columnwidth}
\includegraphics[height=60px, width=60px]{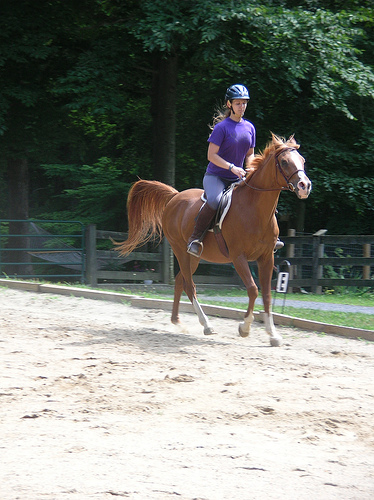}
\end{subfigure}
\hfill
\begin{subfigure}[b]{0.2\columnwidth}
\includegraphics[height=60px, width=60px]{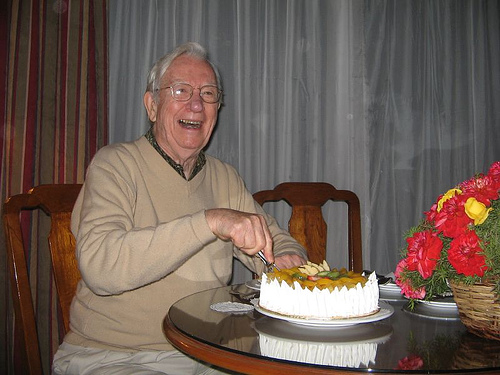}
\end{subfigure}
\hfill
\begin{subfigure}[b]{0.2\columnwidth}
\includegraphics[height=60px, width=60px]{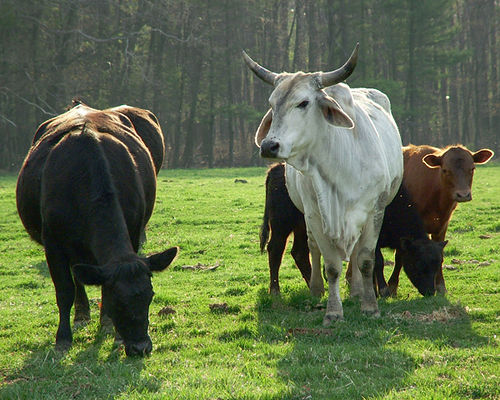}
\end{subfigure}
\hfill
\begin{subfigure}[b]{0.2\columnwidth}
\includegraphics[height=60px, width=60px]{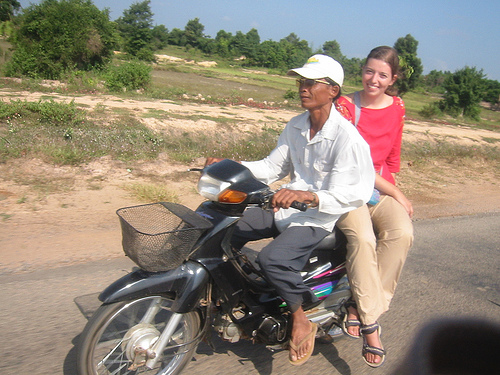}
\end{subfigure}
\vfill
\begin{subfigure}[b]{0.2\columnwidth}
\includegraphics[height=60px, width=60px]{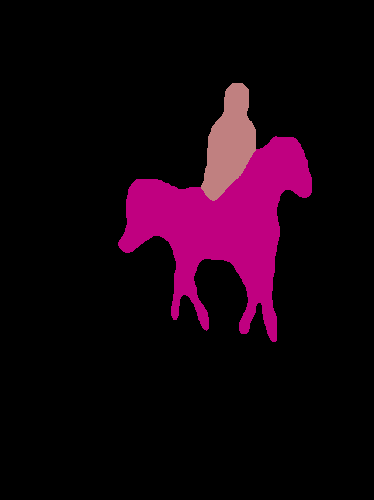}
\end{subfigure}
\hfill
\begin{subfigure}[b]{0.2\columnwidth}
\includegraphics[height=60px, width=60px]{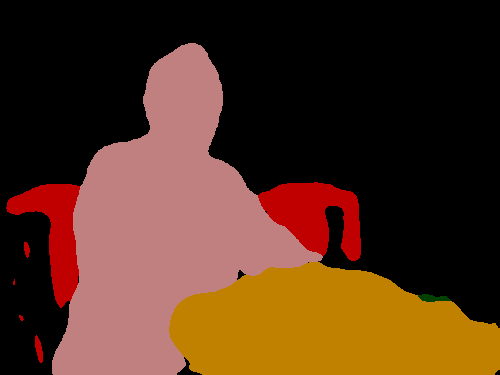}
\end{subfigure}
\hfill
\begin{subfigure}[b]{0.2\columnwidth}
\includegraphics[height=60px, width=60px]{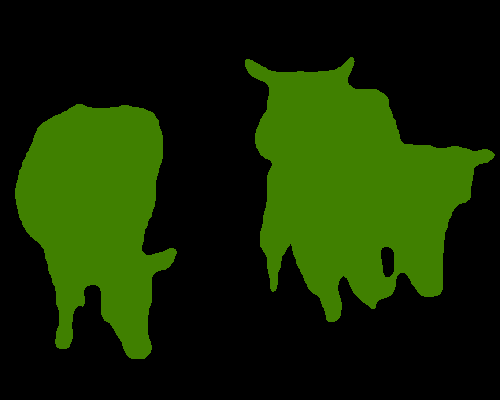}
\end{subfigure}
\hfill
\begin{subfigure}[b]{0.2\columnwidth}
\includegraphics[height=60px, width=60px]{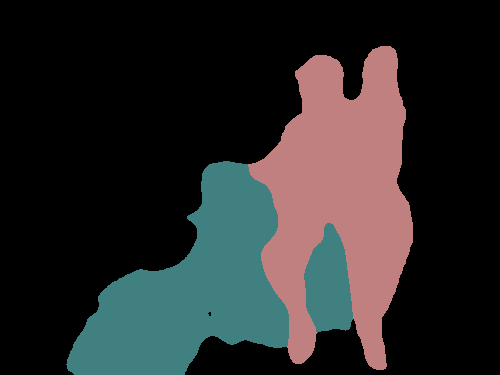}
\end{subfigure}
\setlength{\belowcaptionskip}{-4mm}
\caption{Qualitative results of our method on the PASCAL VOC test set}
\label{fig:pascalResultsQualit}
\end{figure}

\begin{table*}[t!]
\centering
\resizebox{1.5\columnwidth}{!}{
\begin{tabular}{l|ccc|ccc|ccc}
\toprule
  & \multicolumn{3}{|c|}{\textbf{VGG-16}} &  \multicolumn{3}{|c|}{\textbf{ResNet-101}} &   \multicolumn{3}{|c}{\textbf{Ours}}   \\
  \cline{2-10}
 \textbf{Spatial} & \textbf{FLOPS} & \textbf{Parameters} & \textbf{Memory}     & \textbf{FLOPS} & \textbf{Parameters} & \textbf{Memory} & \textbf{FLOPS} & \textbf{Parameters} & \textbf{Memory} \\
 \textbf{Resolution} & \textbf{(in billion)} & \textbf{(in million)} & \textbf{(in MB)}     & \textbf{(in billion)} & \textbf{(in million)} & \textbf{(in MB)} & \textbf{(in billion)} & \textbf{(in million)} & \textbf{(in MB)} \\
\midrule
224                & 1.94  & 0.04   & 26.29      & 0.00  & 0.00   & 0.60   & 0.00  & 0.00   & 0.60   \\
112                & 2.77  & 0.22   & 16.63      & 0.12  & 0.01   & 3.21   & 0.01  & 0.001   & 1.61   \\
56                 & 4.62  & 1.47   & 11.24      & 0.80  & 0.38   & 25.69  & 0.98  & 0.31   & 5.62   \\
28                 & 4.62  & 5.90   & 5.62       & 0.95  & 1.70   & 13.25  & 4.62  & 5.90   & 7.23   \\
14                 & 1.39  & 7.08   & 1.61       & 5.10  & 27.98  & 29.50  & 3.70  & 18.87  & 2.41   \\
7                  & 0.10  & 102.76 & 0.10       & 0.60  & 12.32  & 1.71   & 3.70  & 75.50  & 1.20   \\
\midrule
Overall & 15.45 & 117.43 & 35.19      & \textbf{7.57}  & \textbf{42.39}  & 73.36  & 13.02 & 100.58 & \textbf{18.06} \\
\bottomrule
\end{tabular}
}
\caption{Spatial resolution-wise comparison between VGG-16, ResNet-101, and our custom classification network in terms of FLOPS, number of parameters, and memory.}
\label{tab:compVRO}
\end{table*}

\vspace{-4mm}
\paragraph{Comparison with state-of-the-art methods:}  Following the common convention, we augment the training data with additional annotated images provided in \cite{hariharan2011semantic} and MS-COCO dataset \cite{lin2014microsoft}. Table \ref{tab:pascal} compares the performance of the proposed method with the state-of-the-art methods. Our method achieves mean region intersection over union (mIOU) score of 81.01, which is comparable to the most accurate networks on this dataset. Note that our method delivers competitive accuracy and out-performing previous state-of-the-art methods in terms of inference speed. To the best of our knowledge, our network is the fastest and accurate network on the PASCAL VOC dataset. Segmentation results on the PASCAL VOC 2012 test set are shown in Figure \ref{fig:pascalResultsQualit}. These results suggest that our method has good segmentation properties.

Further, our method can run at different image resolutions while providing an easy trade-off between speed and accuracy. At an image resolution of $224\times 224$, our method performs as good as FCN-8s \cite{shelhamer2017fully}, but runs at 60 fps. We would like to remind the readers that FCN-8s delivers mIOU score of 67.2 when trained at a full image resolution, while our method achieves the same accuracy at almost half of the original image resolution.

\vspace{-4mm}
\paragraph{Speed:} Existing networks use VGG-16 \cite{simonyan2014very} and ResNet-101 \cite{he2016deep} as base feature extractors. These feature extractors are either wide or deep and therefore, they are slow (Table \ref{tab:compVRO}). We exploited the recent advancements in hardware technology (e.g. TitanX can execute tera FLOPS) to make our network fast. Our network executes two convolutional kernels simultaneously in the same time as opposed to a single convolutional kernel in VGG-16 and ResNet-101. Our network has a depth of 18 and is almost $1.5\times$ and $4\times$ faster than VGG-16 and ResNet-101, respectively, while delivering competitive accuracy. A custom and efficient base feature extractor along with a light-weight decoder makes our network fast. 

Furthermore, in contrast to computationally expensive post-processing methods such as CRF, we used a novel residual connection IARC that improved the accuracy without drastically reducing inference speed. 

\section{Results on the ImageNet Dataset}
\label{sec:imageNet}
\begin{figure}[h!]
\centering
\includegraphics[width=0.9\columnwidth]{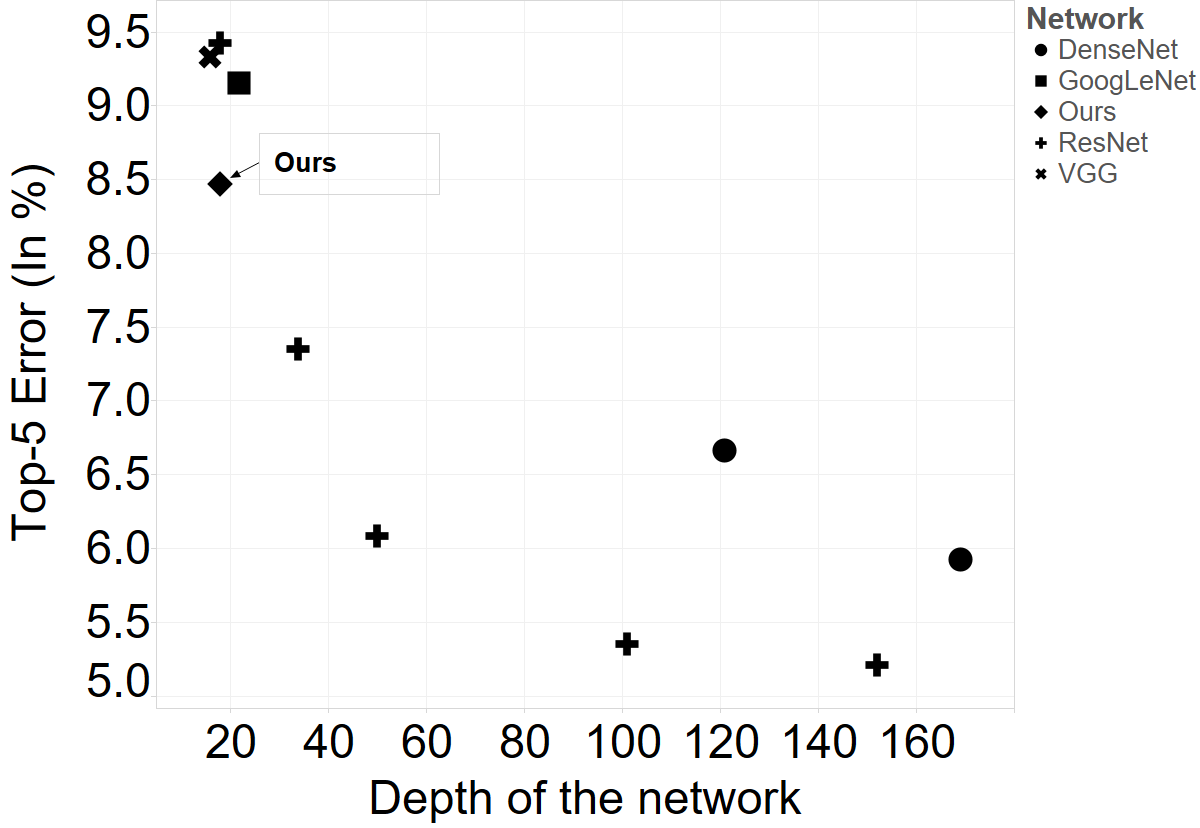}
\caption{Comparison of top-5 error rates (\%, 10-crop testing) on the ImageNet validation set. Among existing networks, our network attains the least error at the similar depth-level (16 to 22). We compared following models: ResNet \cite{he2016deep}, VGG \cite{simonyan2014very}, DenseNet \cite{huang2016densely}, and GoogLeNet \cite{szegedy2015going}.}
\label{fig:imageNetCompar}
\end{figure}

We trained our custom classification network on the ImageNet dataset using the same training strategy as in \cite{he2016deep}. Figure \ref{fig:imageNetCompar} compares the performance of our method with state-of-the-art methods on the ImageNet classification task. Our method attained the least top-5 error when compared with the networks at the similar depth-level. 

For the sake of comparison, we have included the results of very deep networks in Figure \ref{fig:imageNetCompar}, say ResNet-101. The depth of the network has a direct impact on the inference speed. For example, ResNet-101 is $5\times$ slower than ResNet-18. Therefore, we restricted our network to a depth of 18 on the ImageNet dataset. However, our studies on the Cifar dataset suggest that our network can achieve lower error rates with the increase in depth of the network (see Section \ref{sec:cifar}). 

\section{Results on the Cifar Dataset} 
\label{sec:cifar}
Following the previous work (e.g. \cite{he2016deep}), we examine the behavior of our network on the Cifar dataset \cite{krizhevsky2009learning} that consists of 50k training images and 10k test images. The focus of our experiments on the Cifar dataset is to study the impact of scale on the performance of the convolutional neural networks and not pushing the state-of-the-art results. Thus, we construct simple networks using the proposed block and study the impact of depth and width on the performance of the network. 
    
The network takes an input of dimension $32 \times 32$. The first layer is a $3\times3$ convolutional layer, followed by $3$ multi-scale encoding blocks (as shown in Figure 7 in the paper), each with $3n$ layers. Here, $n$ denotes the number of times multi-scale block is repeated. The numbers of filters in these blocks are \{$2F$, $4F$, $8F$\}, where $F$ is the number of the filters in the first convolutional layer. The network ends with an average pooling layer, 10-way (or 100-way)  fully connected layer, and softmax. The \textit{depth} of the network is $\mathbf{9n +2}$.

 The classification error on the Cifar-10 and the Cifar-100 datasets at different depth and width settings is reported in Table \ref{tab:cifarCompare}. The classification error reduces as we make the network deeper and wider. Further, Table \ref{tab:cifarCompare} compares the proposed networks with the state-of-the-art networks. The proposed network is capable of achieving a similar performance to that achieved by very deep networks, but at lesser depth and width. For example, ResNet \cite{he2016deep} with pre-activation \cite{he2016identity} achieves an error rate of 4.92 on the Cifar-10 dataset at a depth of 1001, while the proposed network achieves similar performance (error = 4.99\%) at a depth of 38. Further, the proposed method outperforms WRN \cite{zagoruyko2016wide} at a similar width and depth level. This indicates that aggregating contextual information at different scales is an important aspect separate from the depth and width of the network. 
 
\textbf{Standard vs. Dilated Convolution:} Table~\ref{tab:cifarCompare1}  also reports the result of our network when dilated convolutional filters are used instead of standard convolutional filters. Networks with dilated filters have fewer parameters in comparison to the networks with standard convolutional filters, however, their performance is slightly less than the performance obtained with standard convolutional filters. 

\begin{table}[t!]
    \centering
    \resizebox{\columnwidth}{!}{%
    \small
    \begin{tabular}{|l|c|c|c|c|}
        \hline
           \textbf{Network} & \textbf{Depth} & \multicolumn{1}{|c|}{\textbf{\# Params}} & \multicolumn{1}{|c|}{\textbf{Cifar-10}} & \multicolumn{1}{|c|}{\textbf{Cifar-100}}\\
           \hline
           \multirow{2}{*}{ ResNet \cite{he2016deep}}  & 110 &  \multicolumn{1}{|c|}{ 1.7M} &  \multicolumn{1}{|c|}{ 6.43} & \multicolumn{1}{|c|}{ 25.16}\\
           \cline{2-5}
           & 1202 & \multicolumn{1}{|c|}{10.2M} & \multicolumn{1}{|c|}{7.93} & \multicolumn{1}{|c|}{27.82}\\
           \hline
           \multirow{2}{*}{stoc-depth \cite{huang2016deep}} &  110 & 1.7M &  5.23 &  24.58\\
           \cline{2-5}
           & 1202 & \multicolumn{1}{|c|}{10.2M} & \multicolumn{1}{|c|}{4.91} & \multicolumn{1}{|c|}{--}\\
           \hline
           \multirow{2}{*}{ pre-act \cite{he2016identity}}  & 110 &  1.7M &  6.37 &  --\\
           \cline{2-5}
            & 1001 & \multicolumn{1}{|c|}{10.2M} & \multicolumn{1}{|c|}{4.92} & \multicolumn{1}{|c|}{22.71}\\
           \hline
           \multirow{2}{*}{WRN \cite{zagoruyko2016wide}} &   40 &  \multicolumn{1}{|c|}{ 2.2M} &  \multicolumn{1}{|c|}{ 5.33} &  \multicolumn{1}{|c|}{ 26.04}\\
           \cline{2-5}
            &  28 & \multicolumn{1}{|c|}{36.5M} & \multicolumn{1}{|c|}{4.17} & \multicolumn{1}{|c|}{20.5}\\  
           \hline
           DenseNet \cite{huang2016densely} &  190 &  \multicolumn{1}{|c|}{ 27.2M} &  \multicolumn{1}{|c|}{ \textbf{3.46}} &  \multicolumn{1}{|c|}{ \textbf{17.18}}\\
           \hline        
          \multirow{3}{*}{Ours with standard convolutions ($F=16$)} & 11 & \multicolumn{1}{|c|}{0.39M} & \multicolumn{1}{|c|}{9.8} & \multicolumn{1}{|c|}{34.58} \\
         \cline{2-5} 
          &  20 & \multicolumn{1}{|c|}{0.8M} & \multicolumn{1}{|c|}{7.0} & \multicolumn{1}{|c|}{28.05}\\
          \cline{2-5} 
           &  38 & \multicolumn{1}{|c|}{ 1.62M} & \multicolumn{1}{|c|}{ 6.04} & \multicolumn{1}{|c|}{ 27.06}\\
          \hline
         \multirow{3}{*}{Ours with standard convolutions ($F=32$)}&  11 & 1.57M & 6.72 & 26.85 \\
          \cline{2-5} 
          &   20 & 3.2M & 5.50 & 24.07 \\
          \cline{2-5} 
          & 38 & 6.4M & 4.99  & 23.86\\
          \hline
    \end{tabular}
    }
    \setlength{\belowcaptionskip}{-2mm}
    \caption{\small{Impact of \textit{depth}, \textit{width}, and \textit{scale} on the classification error (in \%) on the Cifar dataset. Here, F represents the width of first residual block. See text for more details.}}
    \label{tab:cifarCompare}
\end{table}

\begin{table}[t!]
    \centering
    \resizebox{\columnwidth}{!}{%
    \small
    \begin{tabular}{|l|c|c|c|c|}
        \hline
           \textbf{Convolution Type} & \textbf{Depth} & \multicolumn{1}{|c|}{\textbf{\# Params}} & \multicolumn{1}{|c|}{\textbf{Cifar-10}} & \multicolumn{1}{|c|}{\textbf{Cifar-100}}\\
          \hline
         \multirow{3}{*}{Standard}&  11 & 1.57M & 6.72 & 26.85 \\
          \cline{2-5} 
          &   20 & 3.2M & 5.50 & 24.07 \\
          \cline{2-5} 
          & 38 & 6.4M & \textbf{4.99}  & \textbf{23.86}\\
          \hline 
          \multirow{3}{*}{Dilated} & 11 & 0.8M & 8.19 & 29.97 \\
          \cline{2-5} 
          &  20 & 1.81M & 5.40 & 24.42 \\
          \cline{2-5} 
          & 38 & 3.71M & 5.38 & 24.12 \\
           \hline
    \end{tabular}
    }
    \setlength{\belowcaptionskip}{-4mm}
    \caption{\small{Impact of standard vs normal convolutions on the classification error (in \%). We replaced the $5\times5$ convolution with a $3\times3$ dilated convolution with dilation rate of 2, so that it has the same effective receptive field as $5\times5$ normal convolution. For these experiments, we set the value of $F=32$. Note that $3\times3$ dilated convolution with a dilation rate of 1 is the same as the standard $3\times3$ convolution.}}
    \label{tab:cifarCompare1}
\end{table}

{\small
\bibliographystyle{ieee}
\bibliography{main}
}

\end{document}